\documentclass[10pt,twocolumn,letterpaper]{article}

\usepackage[pagenumbers]{iccv} 
\usepackage{multirow}
\usepackage{algorithmic}
\usepackage{algorithm}
\usepackage{array}
\usepackage{textcomp}
\usepackage{stfloats}
\usepackage{url}
\usepackage{verbatim}
\usepackage{colortbl}

\usepackage{pifont}
\usepackage{multirow}
\usepackage{makecell}   
\usepackage{amsfonts}   
\usepackage{amsmath}    
\usepackage{booktabs}   
\usepackage{bm}         
\usepackage{setspace}
\usepackage{tikz}
\usepackage{pgfplots}
\usepackage{amssymb}
\usepackage{adjustbox}
\usepackage{float}
\usepackage{subcaption}
\usepackage{graphicx}
\definecolor{iccvblue}{rgb}{0.21,0.49,0.74}
\usepackage[pagebackref,breaklinks,colorlinks,allcolors=iccvblue]{hyperref}

\def\paperID{7028} 
\def\confName{ICCV}
\def\confYear{2025}

\title{\vspace{-1.5 em}MGStream: Motion-aware 3D Gaussian \\for Streamable Dynamic Scene Reconstruction\vspace{-0.2 em}}

\author{
Zhenyu Bao$^{1,2}$ \quad
Qing Li$^{2}$ \quad
Guibiao Liao$^{1,2}$ \quad
Zhongyuan Zhao$^{1,2}$ \quad
Kanglin Liu$^{2\dagger}$ \vspace{0.1 em} \and 
$^1$Peking University \quad
$^2$Pengcheng Laboratory
}

\begin{document}
\twocolumn[{%
	\renewcommand\twocolumn[1][]{#1}%
        \vspace{-1. em}
	\maketitle
        \vspace{-1. em}
	\includegraphics[width=\linewidth]{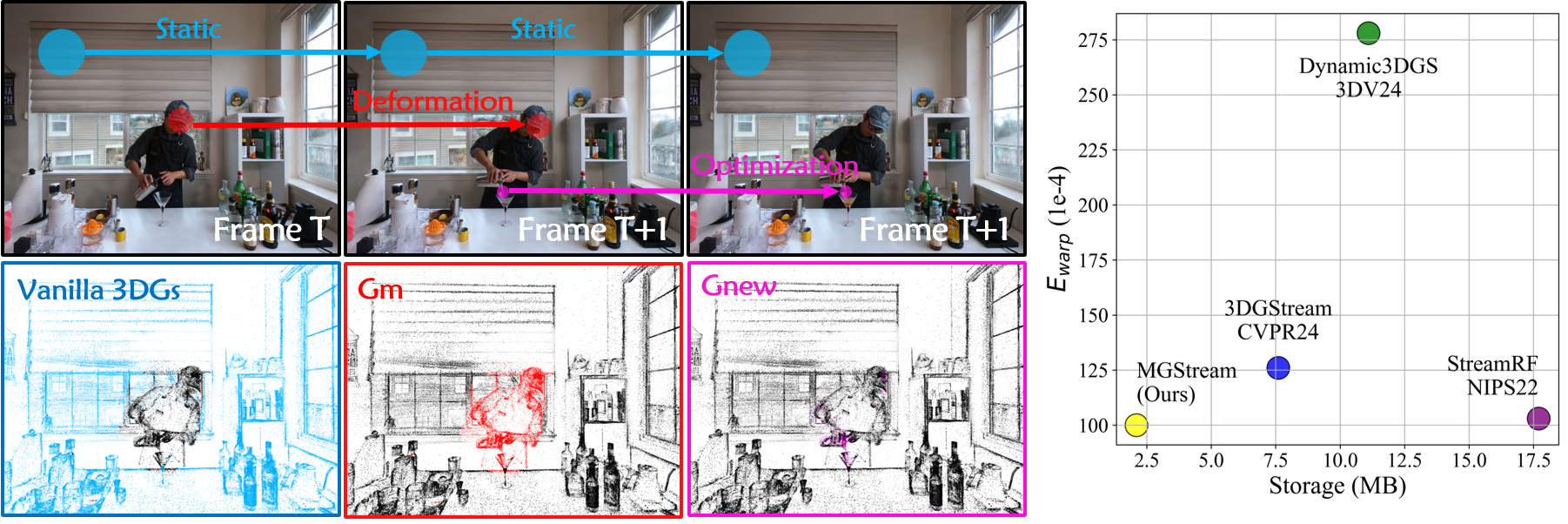}
	\captionof{figure}{MGStream aims to conduct  dynamic novel view synthesis from multiple videos using a per-frame training paradigm. By employing the motion-related 3D Gaussians (3DGs) for modeling the dynamic and the vanilla 3DGs for the static, MGStream achieves high-quality rendering with storage efficiency, and avoids the flickering artifacts.}
        \vspace{1.5 em}
	\label{fig:teaser_figure}
}]
\maketitle
\begin{abstract}
3D Gaussian Splatting (3DGS) has gained significant attention in streamable dynamic novel view synthesis (DNVS) for its photorealistic rendering capability and computational efficiency. Despite much progress in improving rendering quality and optimization strategies, 
3DGS-based streamable dynamic scene reconstruction still suffers from flickering artifacts and storage inefficiency, and struggles to model the emerging objects. To tackle this, we introduce MGStream which employs the motion-related 3D Gaussians (3DGs) to reconstruct the dynamic and the vanilla 3DGs for the static. The motion-related 3DGs are implemented according to the motion mask and the clustering-based convex hull algorithm. The rigid deformation is applied to the motion-related 3DGs for modeling the dynamic, and the attention-based optimization on the motion-related 3DGs enables the reconstruction of the emerging objects. As the deformation and optimization are only conducted on the motion-related 3DGs, MGStream avoids flickering artifacts and improves the storage efficiency. Extensive experiments on real-world datasets N3DV and MeetRoom demonstrate that MGStream surpasses existing streaming 3DGS-based approaches in terms of rendering quality, training/storage efficiency and temporal consistency. Our code is available at: https://github.com/pcl3dv/MGStream.

\end{abstract}
\section{Introduction}

Achieving dynamic novel view synthesis (DNVS) from videos captured by a set of known-pose cameras  remains a challenge in computer vision and computer graphic community. 
The potential applications in  AR/XR/VR devices and broadcasting of competitions have attracted much research.
Conventional methods either rely on geometry to reconstruct the dynamic graphic primitives \cite{tradition_geometry_1,tradition_geometry_2} or utilize image interpolation to obtain novel view synthesis \cite{tradition_interpolation_1,tradition_interpolation_2,tradition_interpolation_3}.
They struggle to synthesize high-quality outputs with novel views due to the complex geometry and appearance of dynamic scenes.

In recent years, volume rendering-based methods, \textit{e.g.,} Neural Radiance Fields (NeRF) \cite{nerf} and 3D Gaussian Splatting (3DGS) \cite{3dgs},  have gained significant attention in DNVS due to their representation capability and high-quality rendering. 
NeRF-based methods fail to achieve efficiency training and real-time rendering, preventing wide applications. As a comparison, 3DGS enables the instant synthesis of novel views with just minutes of training. 
To model the dynamic scenes, three types of methods are widely used: warp-based methods, 4D spatio-temporal methods and per-frame learning methods. 
Specifically, the warp-based methods capture the motions by conducting rigid deformation utilising MLPs \cite{Hyperreel,hypernerf,nerfies,dnerf,deformation3dgs,4DGaussians}.
Alternatively, 4D spatio-temporal methods aim to model the underlying 4D volume across space and time, enabling flexible view synthesis at arbitrary temporal instances \cite{4dgs,4drotorgs,spacetimegs,longvideogs}. Nonetheless, both warp-based and 4D spatio-temporal methods require complete video sequences as input.
For the reliance streaming applications, \textit{e.g.}, live streaming, only per-frame causal inputs are available and on-the-fly training is required, thus preventing from the application of both warp-based and 4D spatio-temporal methods. 
To this end, per-frame learning methods utilize a per-frame training paradigm, which reduces the optimization and training complexity, and enables the online training paradigm. However, typical per-frame learning methods, \textit{e.g.}, 3DGStream \cite{3dgstream} and  Dynamic3DGS \cite{dynamic3dgs}, model the dynamic by conducting deformation on all the 3DGs, leading to storage inefficiency.
In addition, those 3DGs responsible for the reconstruction of the static scenes have been modified, resulting in the flickering artifacts in the synthesized videos.

In this paper, we introduce MGStream for streamable dynamic scene reconstruction. Specifically, MGStream locates the motion-related 3DGs, and employs them and the vanilla 3DGs for modeling the dynamic and static, respectively. To obtain the motion-related 3DGs, MGStream utilizes the optical flow and temporal difference masks to obtain the motion mask, which indicates the motion in the consecutive frames, and establishes the correspondence between the motion-related 3DGs and the motion map via the  Gaussian ID maps (GIM)  and our proposed clustering-based convex hull algorithm.
Then, the rigid deformation is applied to the motion-related 3DGs for modeling the dynamic, and the attention-based optimization enables the motion-related 3DGs to reconstruct emerging objects. 
Since the deformation and optimization operations are applied only for the motion-related 3DGs, MGStream can improve performance in mitigating the flickering artifacts and increasing the storage efficiency over the SOTA models, \textit{e.g.}, 3DGStream and Dynamic 3DGS.
Experiments on the real-world datasets N3DV \cite{n4dnerf} and MeetRoom \cite{streamrf} have demonstrated the effectiveness of our proposed method. Our contributions are as follows:



\begin{itemize}
    \item We introduce MGStream for streamable dynamic scene reconstruction. MGStream utilizes the motion mask and the proposed clustering-based convex hull algorithm to locate the motion-related 3DGs, alleviating the flickering artifacts and improving the storage efficiency.
    \item MGStream models the dynamic by deforming the motion-related 3DGs, and employs the attention map to allow the motion-related 3DGs to reconstruct the emerging objects and guarantee temporal consistency.
    \item Evaluations conducted on the real-world datasets have shown that the proposed method achieves good performance in terms of storage efficiency, temporal stability, rendering quality, and computational efficiency.
\end{itemize}

\section{Related Work}
\subsection{Static Novel View Synthesis}
Novel view synthesis (NVS) has been a long-standing challenge in computer vision, aiming to reconstruct unseen viewpoints of a scene using images captured from one or multiple camera views.
Traditional methods typically relied on view interpolation \cite{chai2000plenoptic,buehler2001unstructured,davis2012unstructured} or 2D generative models \cite{LIU2020308,huang2017beyond,xu2019view}. They  suffer from either low rendering performance or 3D inconsistency issues. The advent of Neural Radiance Fields (NeRF) makes a significant breakthrough, encoding scene radiance information into neural network parameters and enabling high-fidelity and view-consistent free-viewpoint synthesis \cite{nerf}. 
Various methods have been developed to address inherent limitations of NeRF, including anti-aliasing \cite{mipnerf,zipnerf,badnerf}, low inefficiency \cite{DVGO,Plenoxels,InstantNGP}, insufficient view constraints \cite{sparsenerf,freenerf,regnerf}, and poor geometry reconstruction\cite{neus,volsdf,monosdf}. However, the implicit NeRF representations necessitate costly sampling queries, forcing subsequent methods to grapple with trade-offs between training time, inference speed, storage requirements, and reconstruction quality.
Recently, Kerbl et al. introduce 3D Gaussian Splatting (3DGS) for scene representation and high quality rendering \cite{3dgs}. By combining customized rasterization manner, 3DGS achieves rapid training and real-time rendering performance for large-scale, unbounded scenes while maintaining rendering quality comparable to state-of-the-art NeRF variants such as Mip-NeRF360 \cite{mipnerf360}. The following 3DGS-based researches focus on reducing the memory usage \cite{scaffold-GS,compact3d}, aliasing issues \cite{mipsplatting}, and few-shot view conditions \cite{loopsparsegs,fsgs,dngaussian}. 
The elegant explicit representation and superior real-time rendering capabilities of 3DGS have also inspired further valuable researches, including 3D manipulations \cite{pointmove,gaussianeditor}, and scene understanding \cite{clipgs,langsplat,semanticgs}. 
These methods concentrate on static scenes, and can not directly be applied to dynamic scenes.


\subsection{Dynamic Novel View Synthesis}
DNVS aims to produce free-view videos from multi-view video inputs. Benefiting from the high rendering quality of the radiance field, various approaches have been proposed and can primarily be grouped into three categories: warp-based methods, 4D spatio-temporal model-based methods,  and per-frame learning-based methods.


\textbf{Warp-based Methods.}
Warp-based methods model the dynamic scene by estimating the transformation between the current and the canonical radiance field generated from multi-view images.
Pumarola et al. construct a static canonical space and use an MLP to query temporal deformations \cite{dnerf}. 
Yang et al. achieve real-time rendering by querying Gaussian primitive changes with an annealing mechanism for smooth transitions \cite{deformation3dgs}. 
Similarly, 4DGaussians combines k-planes with 3D Gaussian primitives, enabling efficient training and rendering of dynamic scenes by querying changes in position, rotation, and scale at arbitrary time \cite{4DGaussians}. E-D3DGS \cite{bae2024per} introduces primitive-level features and coarse-to-fine temporal embeddings to handle various deformations of different 3DGs. While these methods excel at capturing rigid transformations, they struggle with complex topological changes and new objects.

\textbf{4D Spatio-Temporal Models.} 
4D spatio-temporal models achieve dynamic novel view synthesis via incorporating time parameters into the radiance field, \textit{e.g.}, K-Planes decomposes spatio-temporal scenes into six planes, separating temporal and spatial attributes while maintaining continuity and reducing memory usage \cite{kplanes}. 4DGS extends 3D Gaussian Splatting using dual-quaternion 4D rotations to model dynamics \cite{4dgs}, while 4D RotorGS employs rotor-based 4D rotations to improving interpretability \cite{4drotorgs}. SpacetimeGS conducts temporal opacity and parametric motion/rotation for vanilla 3DGS to capture the dynamic \cite{spacetimegs}. Temporal Gaussian Hierarchy employs multi-level 4DG primitives for long video reconstruction \cite{longvideogs}. However, these spatial-temporal approaches often struggle to preserve fine details, compromising rendering quality.


\textbf{Per-frame Learning.}
Different from the wrap-based and 4D spatio-temporal methods, per-frame learning method is capable of streaming.
StreamRF pioneers using Plenoxels with narrow-band tuning for inter-frame differences \cite{streamrf}. ReRF enhances this approach using a custom pooling motion grid with residual feature grids to model long sequence frames efficiently \cite{rerf}. These methods are limited by the inherent trade-offs between rendering quality and inference speed imposed by NeRF-based architectures. Recently, Dynamic3DGS introduces the local rigidity constraints for Gaussian primitives movements, which enabled dense scene tracking \cite{dynamic3dgs}. Sun et al. propose a Neural Transformation Cache (NTC) for modeling frame differences, enabling superfast single-frame training and real-time rendering \cite{3dgstream}.
Despite achieving fast per-frame training (about 10s), the costly static storage requirements and the severe flickering artifacts hinder practical real-world applications. HiCoM \cite{gao2024hicom} retains densified primitives across frames while synchronously removing an equal number of low-occupancy ones to maintain stable streaming. QUEEN \cite{girish2024queen} introduces a quantized sparse framework that flexibly models all Gaussian parameters, achieving lower storage costs and higher rendering quality. However, neither \cite{gao2024hicom, girish2024queen} addresses the issue of flickering in static regions during streaming.
In contrast, MGStream models the dynamic by deforming and optimizing the motion-related 3DGs. Since the deformation is applied for the motion-related 3DGs, MGStream alleviates the flickering artifacts and improves the storage efficiency.
\begin{figure*}[t]

  \centering
  \includegraphics[width=\linewidth]{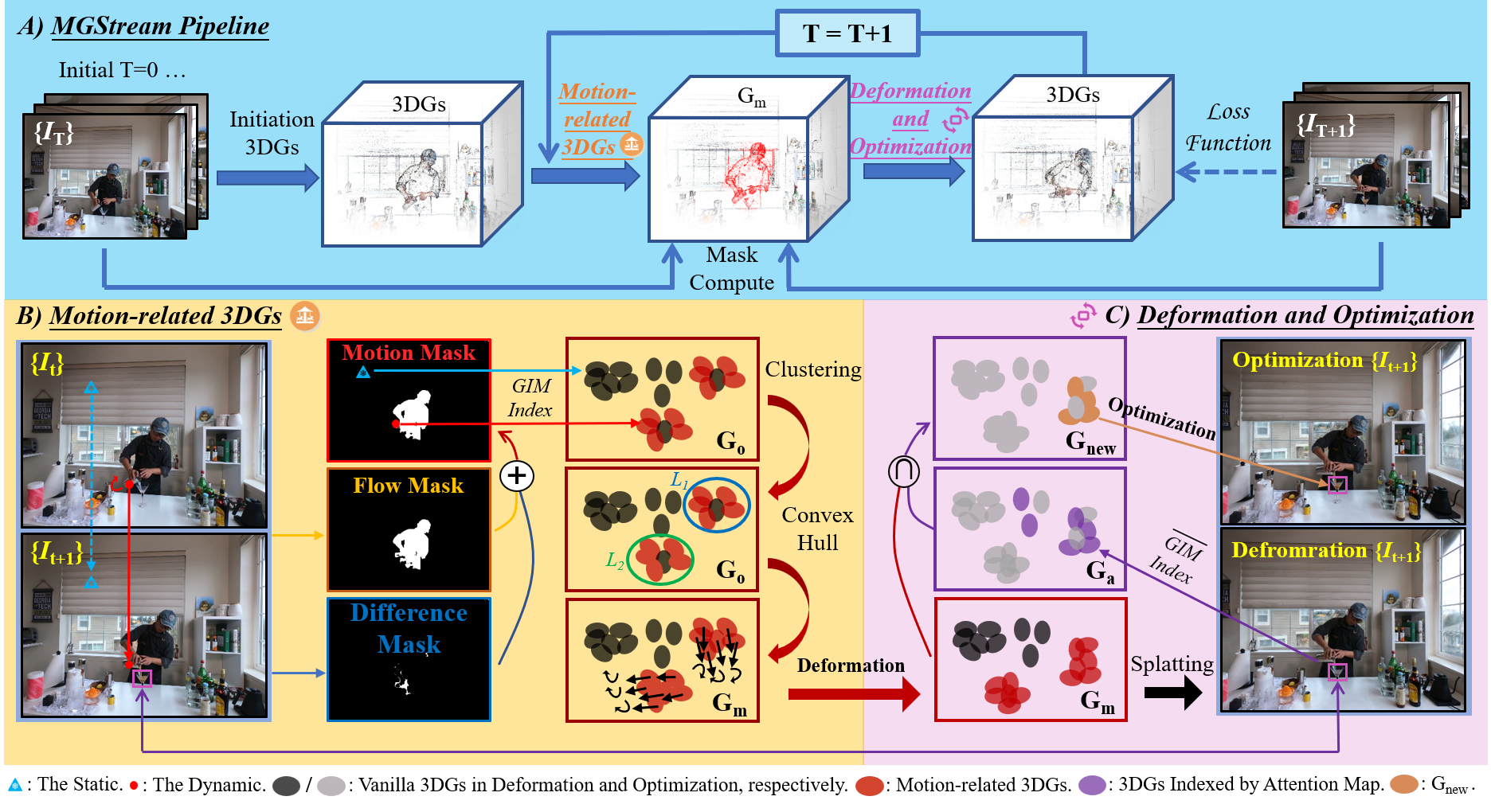}
  \vspace{-1.9em}
   \caption{Overview of the proposed \textbf{MGStream}. 
   \textbf{A)} shows the MGStream pipeline.
   Firstly, MGStream initiates the 3DGs with multi-view inputs at timestep 0. For subsequent frames, MGStream utilizes the 3DGs in the previous timestep as initialization.
   Then, MGStream locates the motion-related 3DGs, which are deformed and optimized for modeling the dynamic.
   \textbf{B)} elaborates the way for finding the motion-related 3DGs. MGStream employs the optical flow and temporal difference of adjacent frames to determine the motion mask, and establish the correspondence between the motion mask and the motion-related 3DGs via the GIM and clustering-based convex hull algorithm.
   \textbf{C)} shows the deformation and optimization on the motion-related 3DGs.
   All the motion-related 3DGs are deformed for modeling the dynamic, and the attention map is used to find those responsible for the emerging objects. Eventually, optimization is applied on those for reconstructing the new objects.
   }
   \vspace{-1em}
   \label{fig:pipeline}
\end{figure*}

\textbf{Concurrent Works.}
SwinGS \cite{SwinGS} integrates spacetime Gaussian with Markov Chain Monte Carlo (MCMC) to reduce transmission costs and model long-time series video sequences, while our method models the dynamic scene mainly relying on the motion-related 3DGs, which mitigate the flickering artifacts and improve the storage efficiency. In a similar way, S4D \cite{s4d} locates the moving Gaussian under the guidance of the user, and DASS \cite{DASS} utilizes optical flow and segmentation models to define the motion properties of 3DGs based on a learning approach.
In contrast, our method identifies the motion-related 3DGs using the motion mask and the proposed clustering-based convex hull algorithm without the learning process.

\section{Preliminaries} \label{Preliminaries} 
3D Gaussian Splatting (3DGS) utilizes discrete primitives for geometry representation. Each 3D Gaussian (3DG) is parameterized by the position $u$, opacity $o$, scale $s$, rotation $q$, and 
color $c$, which is expressed by a set of spherical harmonic coefficients ($sh$s).
 3DGs can be rendered via a differentiable splatting-based rasterizer, and the color $C$ of a pixel is computed by alpha composition:
\begin{equation}
C = \sum_{i =1} \mathbf{c}_i \alpha_i \prod_{j=1}^{i-1} (1 - \alpha_j), 
\label{eq:3dgs}
\end{equation}
where $\alpha$ represents the blending weight which is given by evaluating the 2D projection of the 3DGs multiplied by the opacity $o$. The parameters of each 3DG are  optimized by minimizing the difference $\mathcal{L}_{color}$ between the rendered image $\tilde{I}$ and the ground truth image:  
\begin{equation}
\mathcal{L}_{color} = (1-\lambda)\mathcal{L}_{1}(\tilde{I}, I) + \lambda\mathcal{L}_{D-SSIM}(\tilde{I}, I).
\label{eq:3dgs_loss}
\end{equation}
Following vanilla 3DGS \cite{3dgs}, $\mathcal{L}_{color}$ contains the $\mathcal{L}_{1}$ and $\mathcal{L}_{D-SSIM}$ term, and we also set $\lambda$ to 0.2 in all experiments.

\section{Method}

MGStream aims to reconstruct free-viewpoint videos of dynamic scenes from multi-view video streams using a per-frame training paradigm, and the framework is illustrated in \cref{fig:pipeline}. MGStream utilizes the vanilla 3DGS to initiate the 3DGs with the multi-view inputs at timestep 0. For subsequent timesteps, MGStream employs 3DGs in the previous timestep as an initialization. In order to model the dynamic, previous methods conduct deformation and optimization on all the 3DGs, causing flickering artifacts and storage inefficiency. To tackle this, MGStream models the dynamic scenes mainly through motion-related 3DGs. Specifically, motion-related 3DGs are responsible for modeling the scenes with motions including the emerging objects, while the vanilla 3DGs reconstruct the static. The implementation of the motion-related 3DGs is introduced in \cref{method_motion_related}, and the deformation and optimization are conducted in \cref{method_deformation_optimization}.


\subsection{Motion-related 3DGs} \label{method_motion_related}
Given the current frame $I_t$, the previous frame $I_{t-1}$, and the initialized 3DGs $G_{t-1}$ from the previous frame, MGStream identifies the motion-related 3DGs to model the dynamic. Specifically, we utilize the optical flow and temporal difference of adjacent frame to determine the motion mask, and establish the correspondence between the motion-related 3DGs and 2D motion mask with the Gaussian ID maps of maximum alpha-blending weights (GIM) and the clustering-based convex hull algorithm, as shown in \cref{fig:pipeline}.

\begin{figure}[t!]  
	\centering
	\begin{subfigure}{\linewidth}
            \begin{minipage}[t]{0.31\linewidth}
                \centering
                \includegraphics[width=1\linewidth]{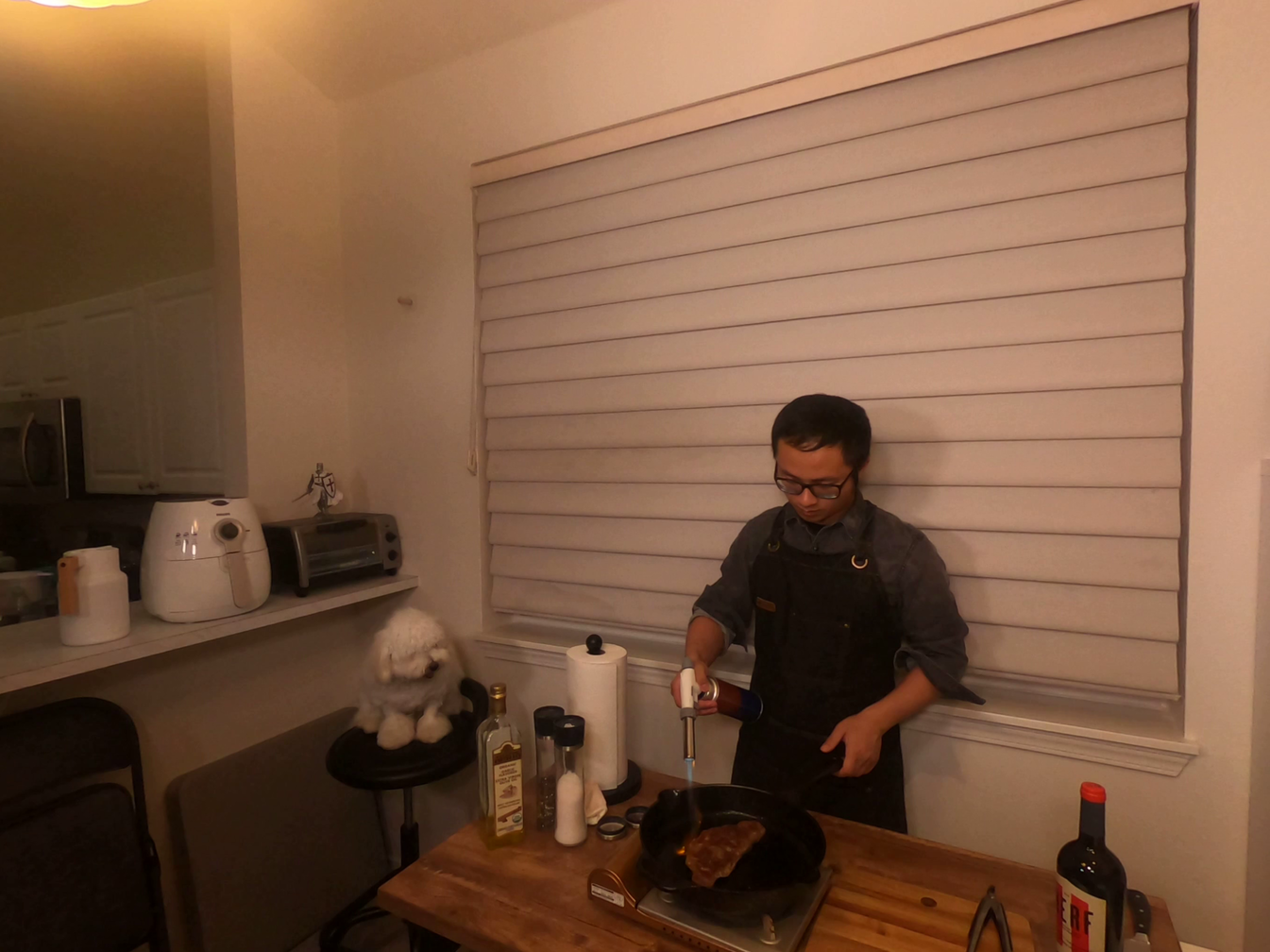}
                \caption{$I_{t-1}$}
            \end{minipage}
            \begin{minipage}[t]{0.31\linewidth}
                \centering
                \includegraphics[width=1\linewidth]{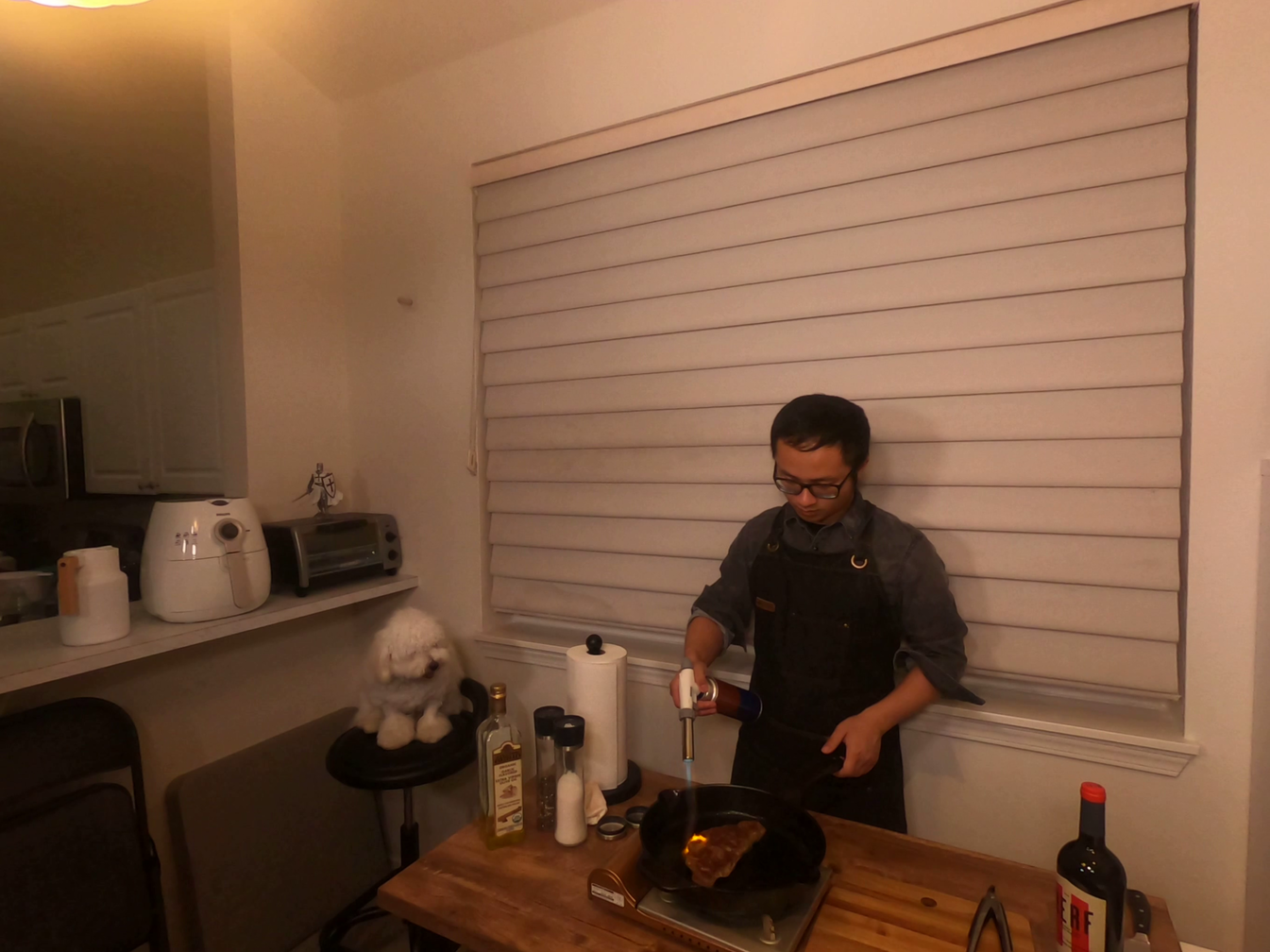}
                \caption{$I_{t}$}
            \end{minipage}
            \begin{minipage}[t]{0.31\linewidth}
                \centering
                \includegraphics[width=1\linewidth]{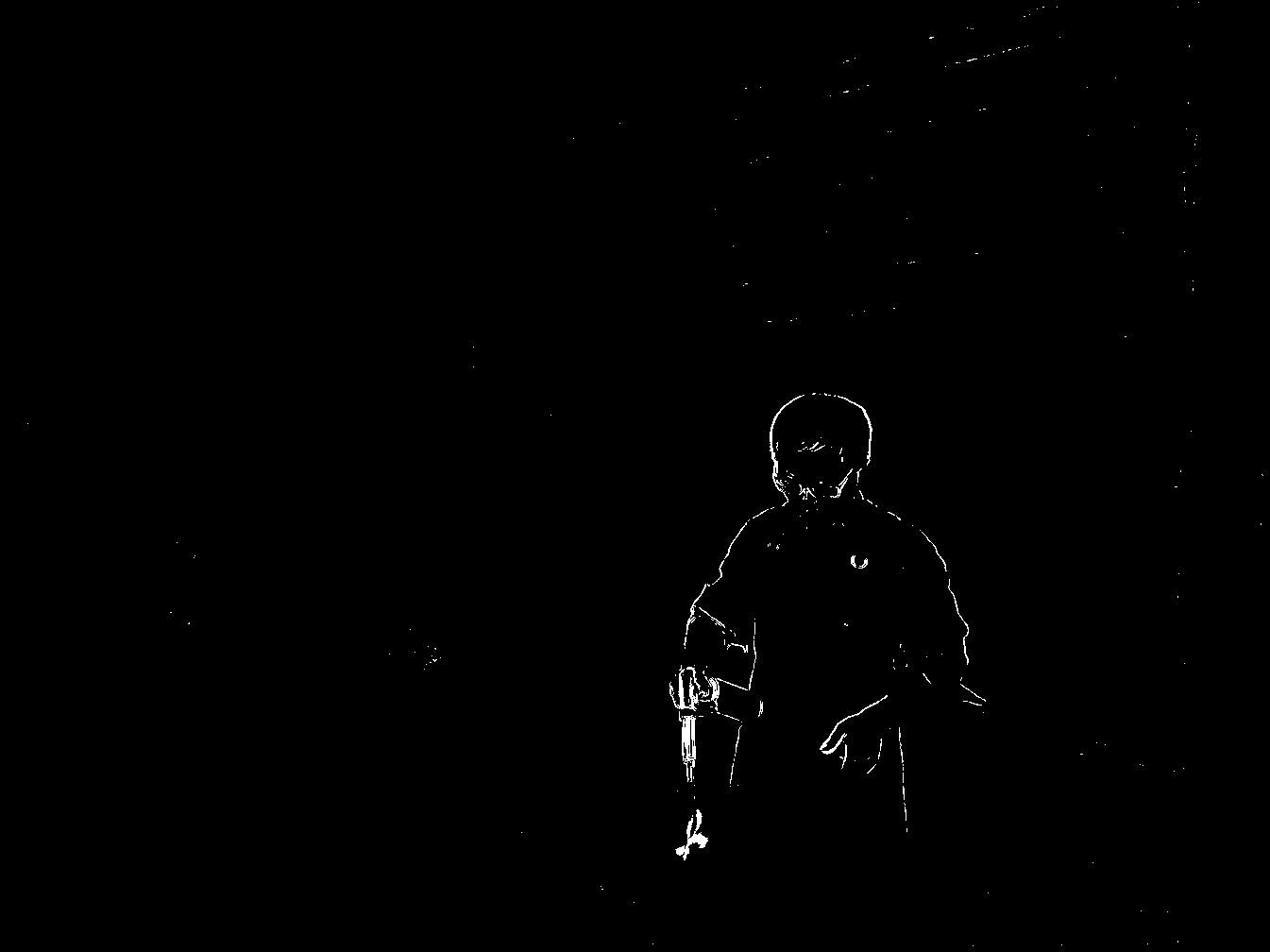}
                \caption{$|I_t-I_{t-1}|$}
            \end{minipage}
        \end{subfigure}

        \vspace{0.1cm}
	\begin{subfigure}{\linewidth}
            \begin{minipage}[t]{0.31\linewidth}
                \centering
                \includegraphics[width=1\linewidth]{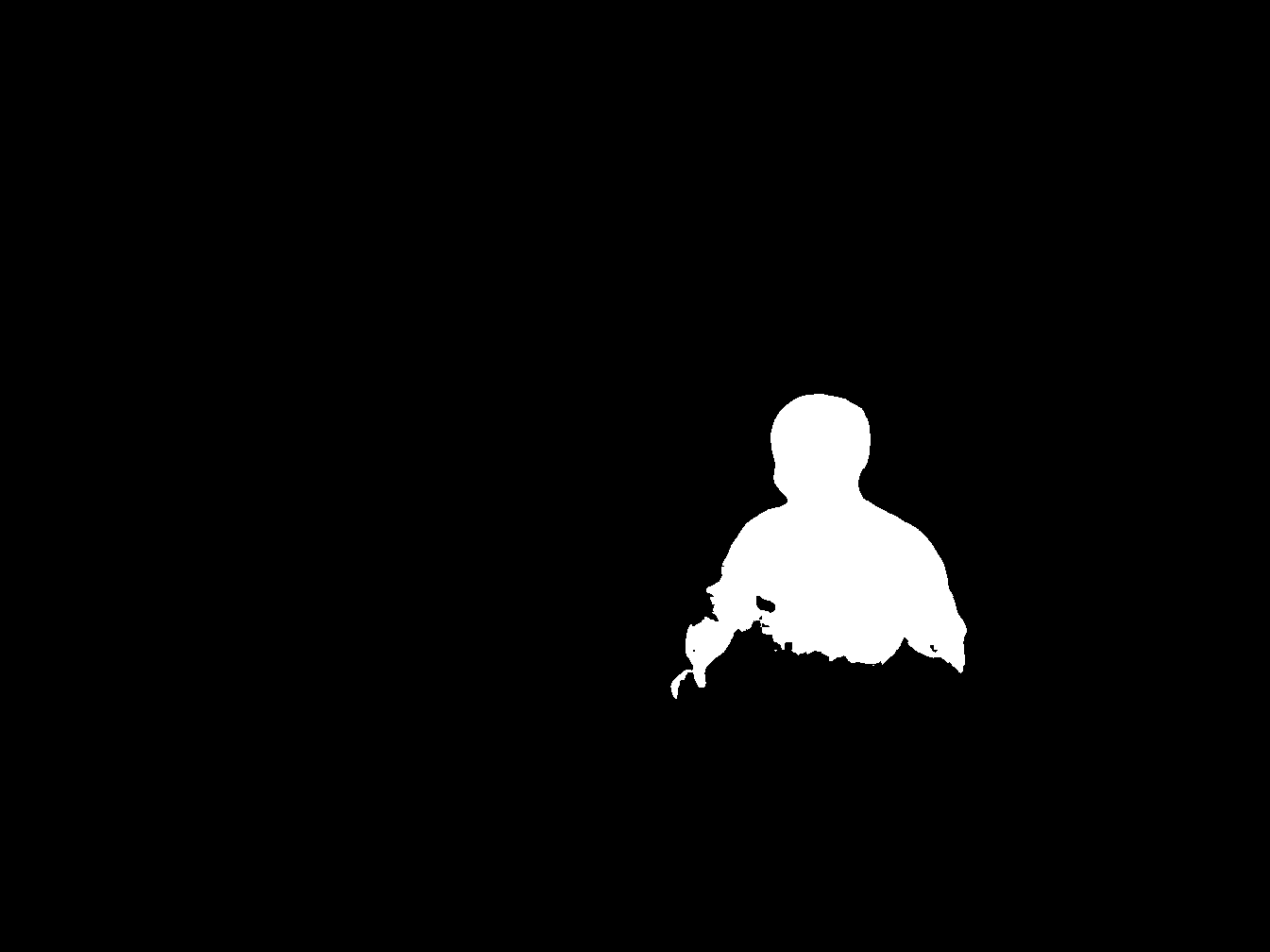}
                 \caption{Flow Mask}
            \end{minipage}
            \begin{minipage}[t]{0.31\linewidth}
                \centering
                \includegraphics[width=1\linewidth]{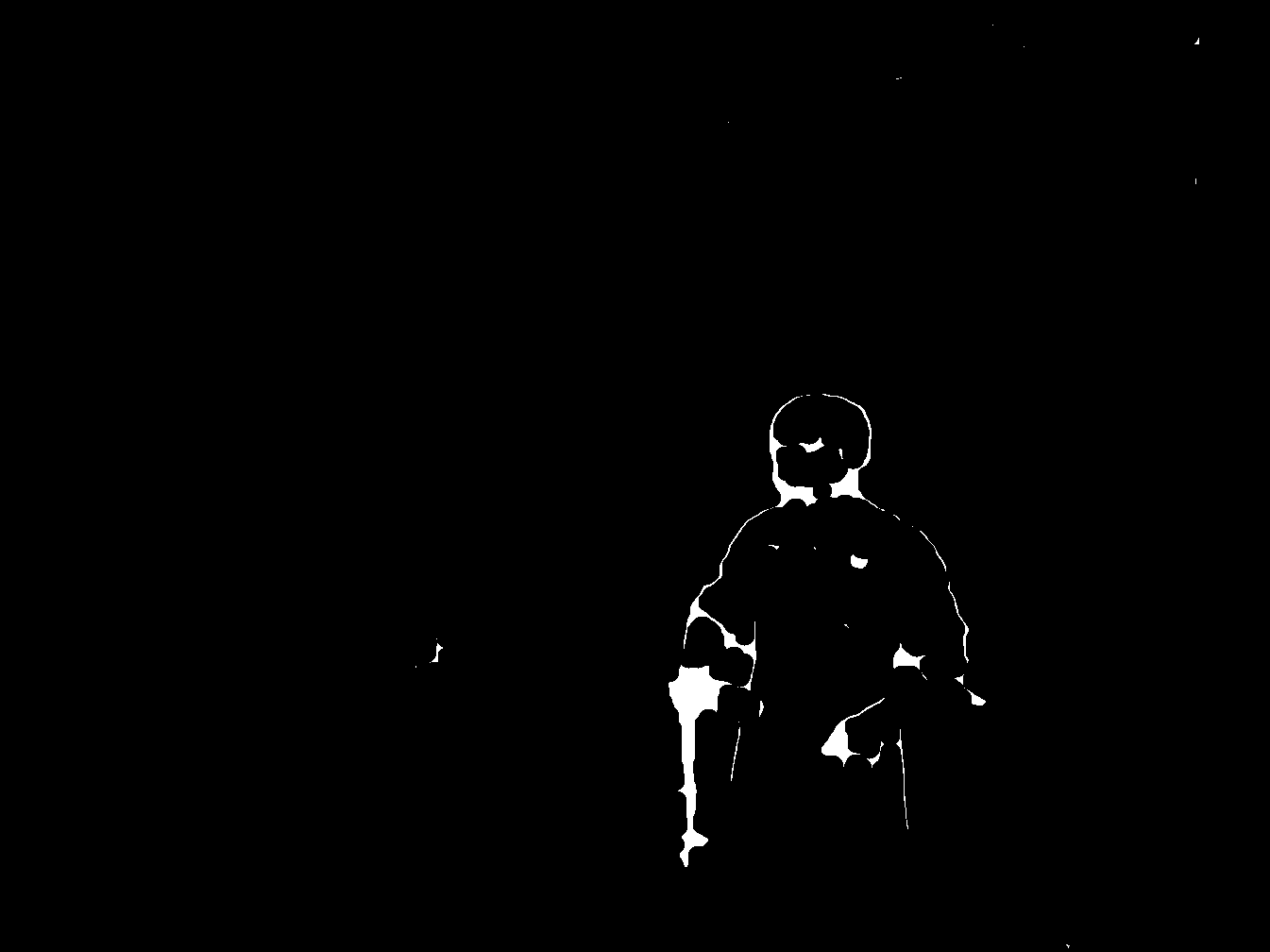}
                \caption{Diff Mask}
            \end{minipage}
            \begin{minipage}[t]{0.31\linewidth}
                \centering
                \includegraphics[width=1\linewidth]{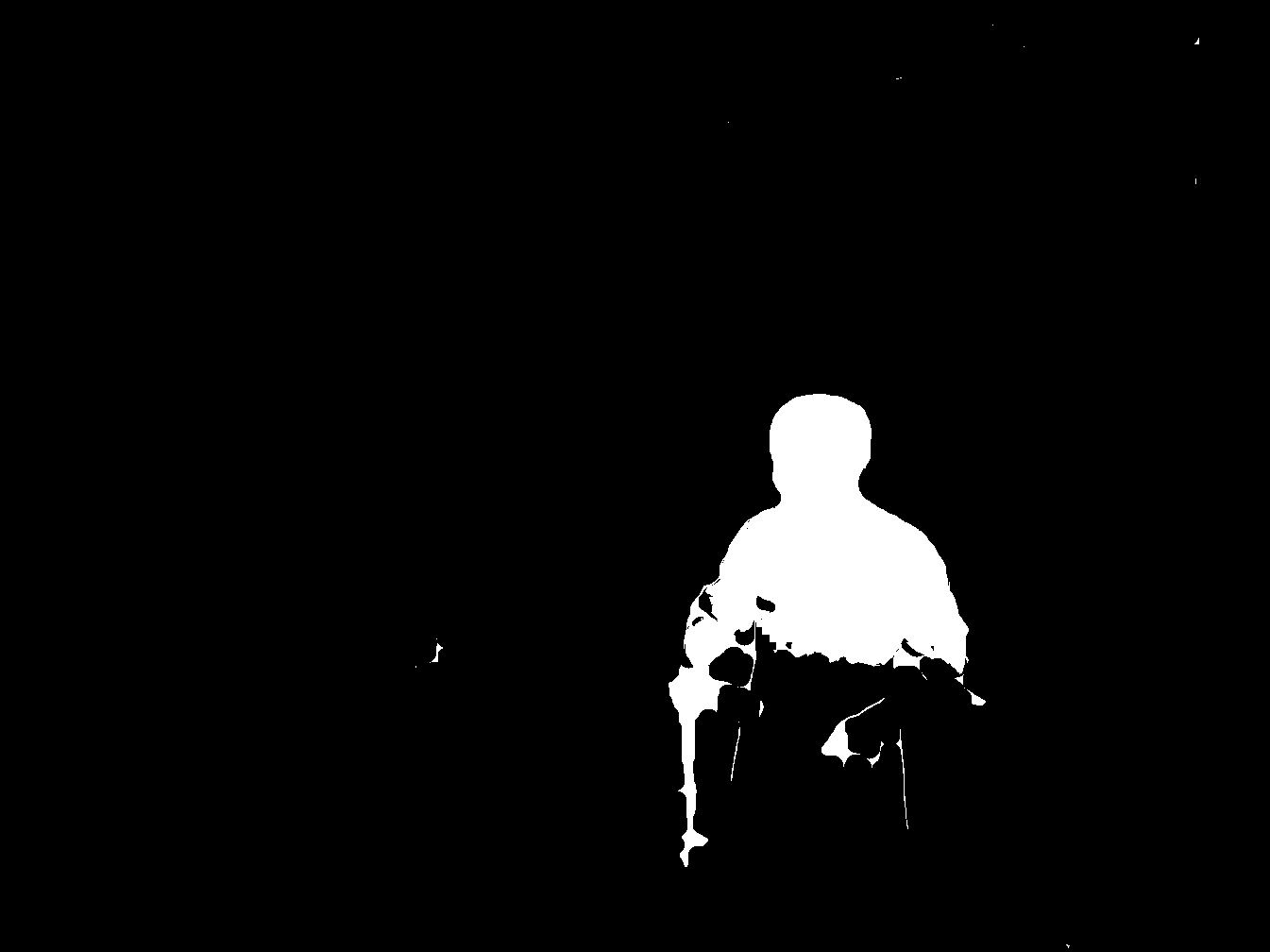}
                \caption{Motion Mask}
            \end{minipage}
        \end{subfigure}

       \caption{Illustration of Motion Mask. (a) and (b) represent images in frame $t-1$ and $t$ from identical camera view, respectively, and (c) is the difference  between (a) and (b), \textit{i.e.}, $|I_t-I_{t-1}|$. (d), (e) and (f) are the optical flow mask, temporal difference mask, and motion mask, respectively.}
       \vspace{-0.8em}
       \label{fig:Motion_Mask}
\end{figure}

\textit{Motion Mask}. MGStream employs the optical flow and temporal difference of adjacent frames to determine the motion mask. 2D optical flow model enables the detection of pixel-level motion velocities between consecutive frames and effectively identifies object motion through predefined threshold \cite{motion_detection_1,motion_detection_2}, contributing to finding the motion-related 3DGs. However, the pre-trained optical flow model can hardly distinguish the tiny motion, \textit{e.g.} the motion of flaming gun is ignored in the obtained optical flow model as shown in \cref{fig:Motion_Mask} (c) and (d). To address this, we employ the temporal difference of adjacent frame as well for improving the accuracy of the motion mask, derived as:
\begin{equation}
       \hat{M} = \delta (\mathcal{O}(I_t,I_{t-1}), \tau ) \circ \mathfrak{E}(\mathfrak{D} (\text{abs}(I_t-I_{t-1}))),
      \label{eq:motion}
\end{equation}
where $\delta$ is the step function, $\mathcal{O}$ is the optical flow model, and $\tau$ is the threshold, which is set to $1$ in our work. $\mathfrak{D}$ and $\mathfrak{E}$ represent the \textit{DILATE} and \textit{ERODE} of morphological processing. The obtained motion mask is shown in \cref{fig:Motion_Mask}.

 \textit{Clustering-based Convex Hull Algorithm}. Establishing the correspondence between the motion mask and the motion-related 3DGs remains a challenging task. Previous methods either employ back-projection of 2D points into 3D space to locate motion-related neighbor 3DGs using depth information \cite{motionawaregs}, or establish correlations between 2D optical flow and 3DG motion velocities \cite{gaussianflow}. However, these methods are computationally intensive, making them unsuitable for on-the-fly per-frame training. 
 To address this, we introduce a clustering-based convex hull algorithm to find the motion-related 3DGs using the motion mask.
 Firstly, Gaussian ID maps of maximum alpha-blending weights (GIM) are utilized to back-projection the pixel in the motion mask to the 3DGs:
 
 \begin{equation}
       G_{o}=\bigcup_{i=1}^{V} (GIM_i \circ \hat{M_i}),
      \label{eq:motion}
\end{equation}
where $V$ is the total camera views and $G_{o}$ is the obtained motion-related 3DGs via back-projection, and GIM represents the index of Gaussians with the highest weight during the alpha blending process: 
\begin{equation}
       \text{GIM}^n=\max(\alpha_i\prod_{j=1}^{i-1} (1-\alpha_j)),
      \label{eq:GIM}
\end{equation}
where $n$ is the pixel index.

\begin{figure}[t!]  
	\centering
	\begin{subfigure}{\linewidth}
            \begin{minipage}[t]{0.31\linewidth}
                \centering
                \includegraphics[width=1\linewidth]{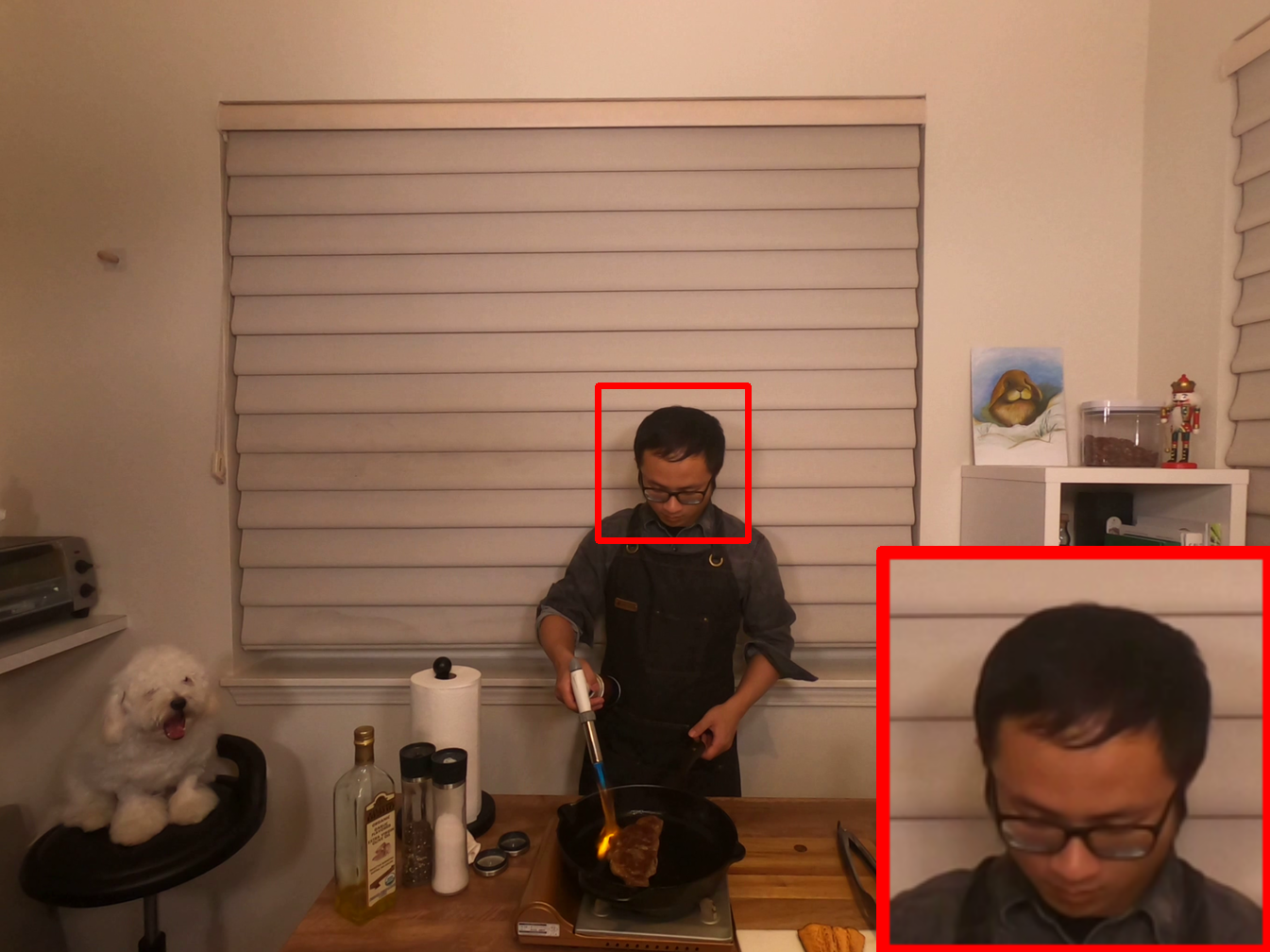}
                \caption{$I_{t-1}$}
            \end{minipage}
            \begin{minipage}[t]{0.31\linewidth}
                \centering
                \includegraphics[width=1\linewidth]{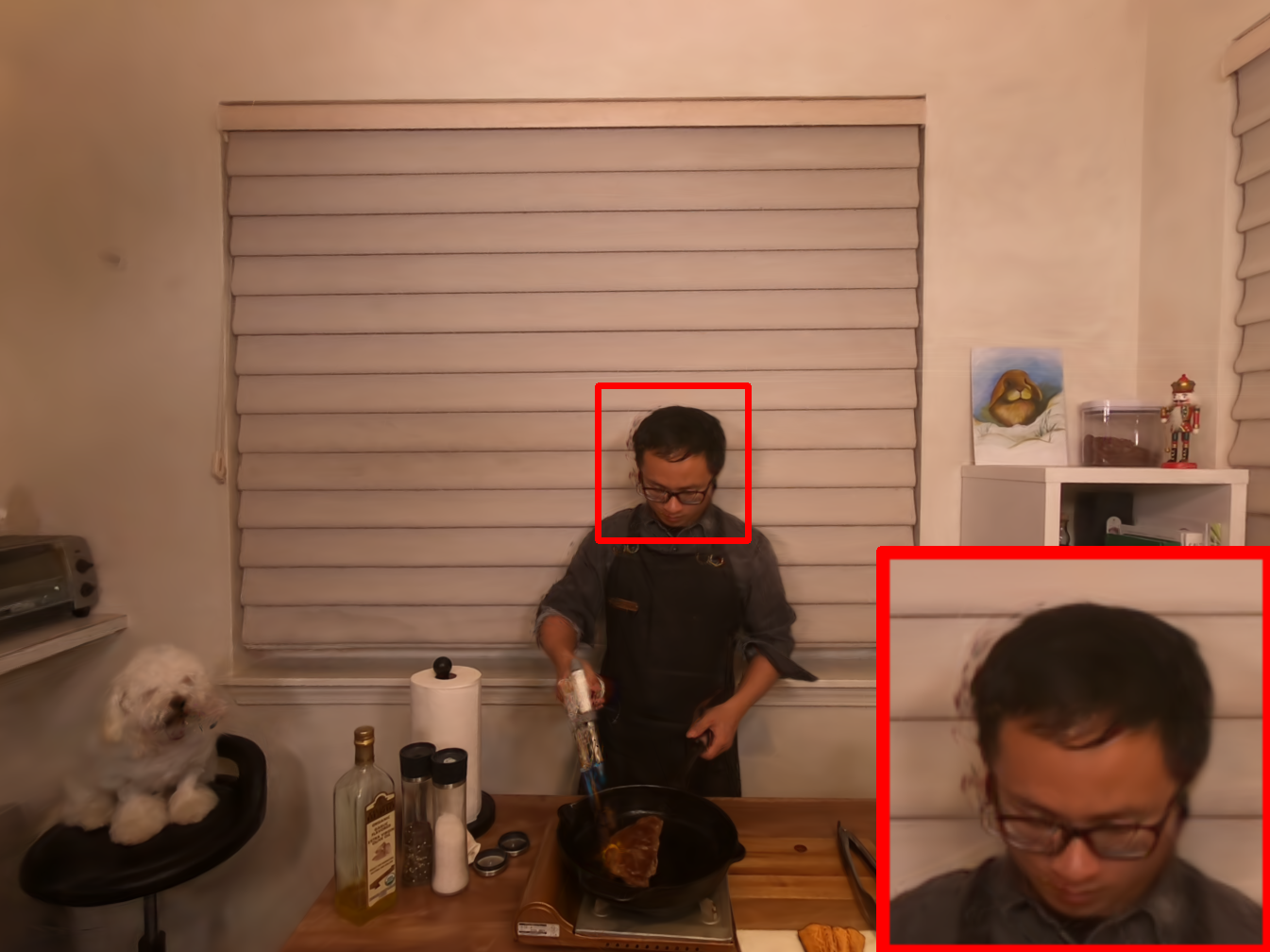}
                \caption{$I_{t}$ w/o Convex Hull}
            \end{minipage}
            \begin{minipage}[t]{0.31\linewidth}
                \centering
                \includegraphics[width=1\linewidth]{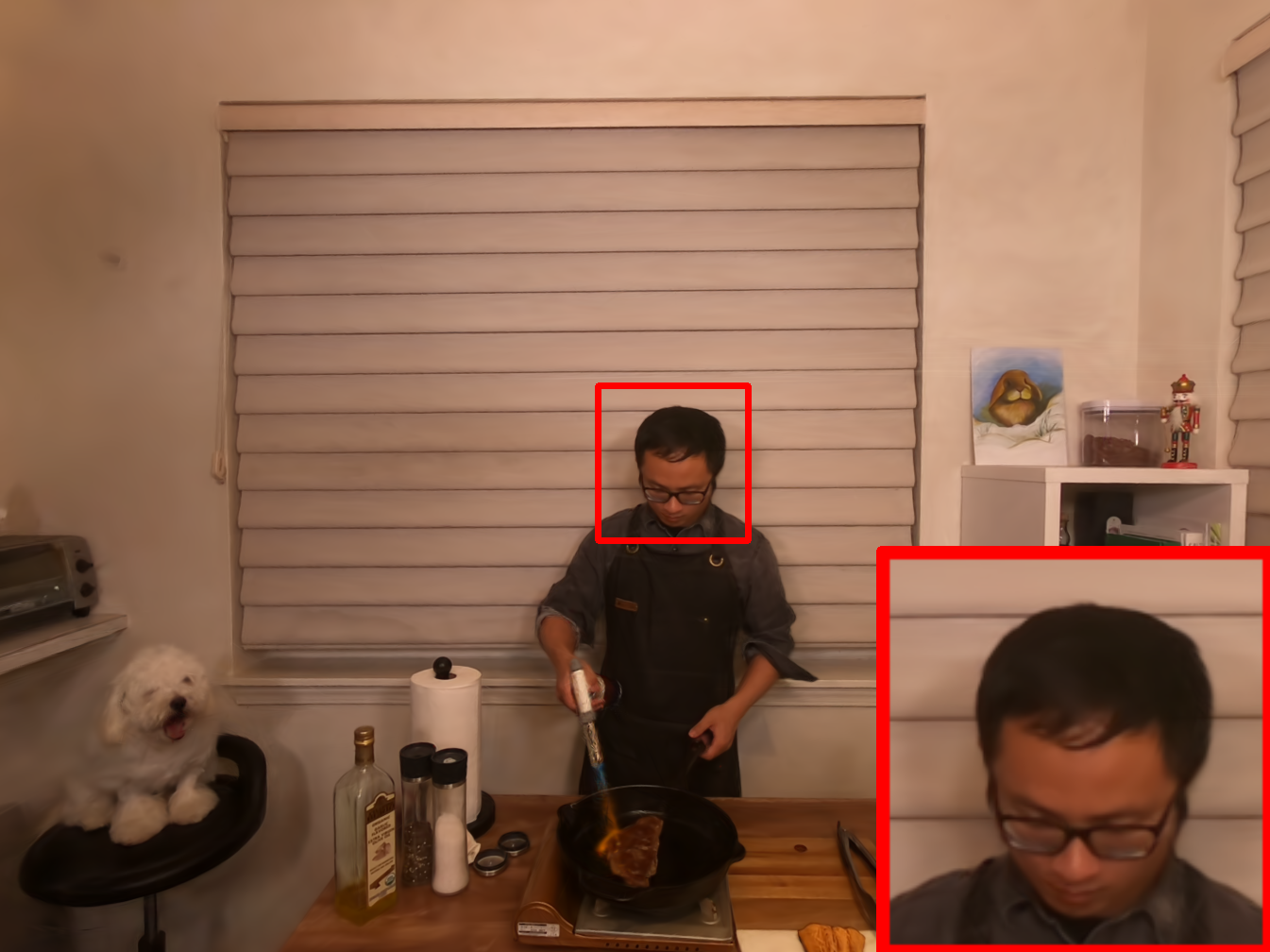}
                \caption{$I_{t}$ w/ Convex Hull}
            \end{minipage}
        \end{subfigure}

        \vspace{0.1cm}
	\begin{subfigure}{\linewidth}
            \begin{minipage}[t]{0.31\linewidth}
                \centering
                \includegraphics[width=1\linewidth]{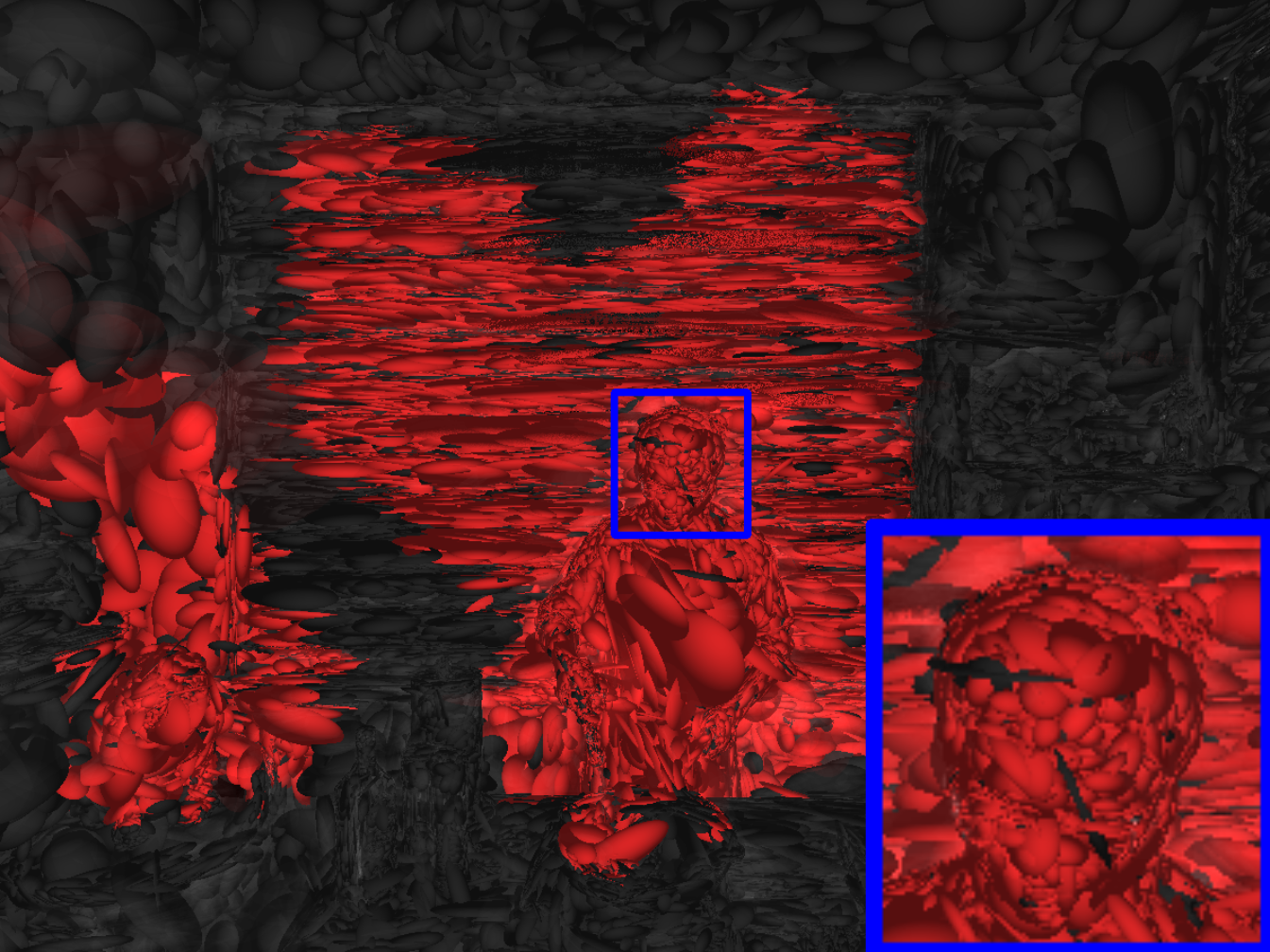}
                 \caption{$G_o$}
            \end{minipage}
            \begin{minipage}[t]{0.31\linewidth}
                \centering
                \includegraphics[width=1\linewidth]{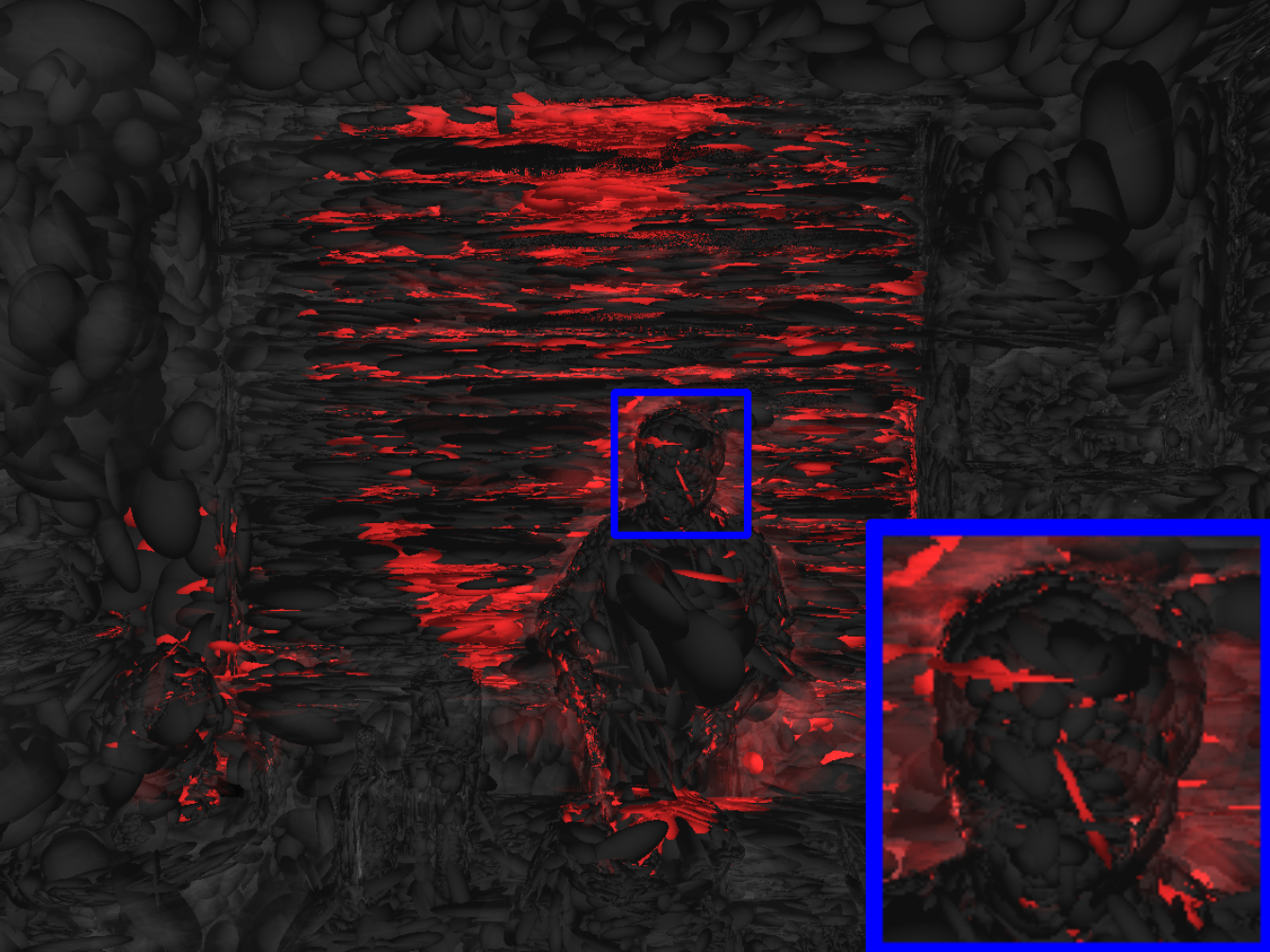}
                \caption{$G_i$}
            \end{minipage}
            \begin{minipage}[t]{0.31\linewidth}
                \centering
                \includegraphics[width=1\linewidth]{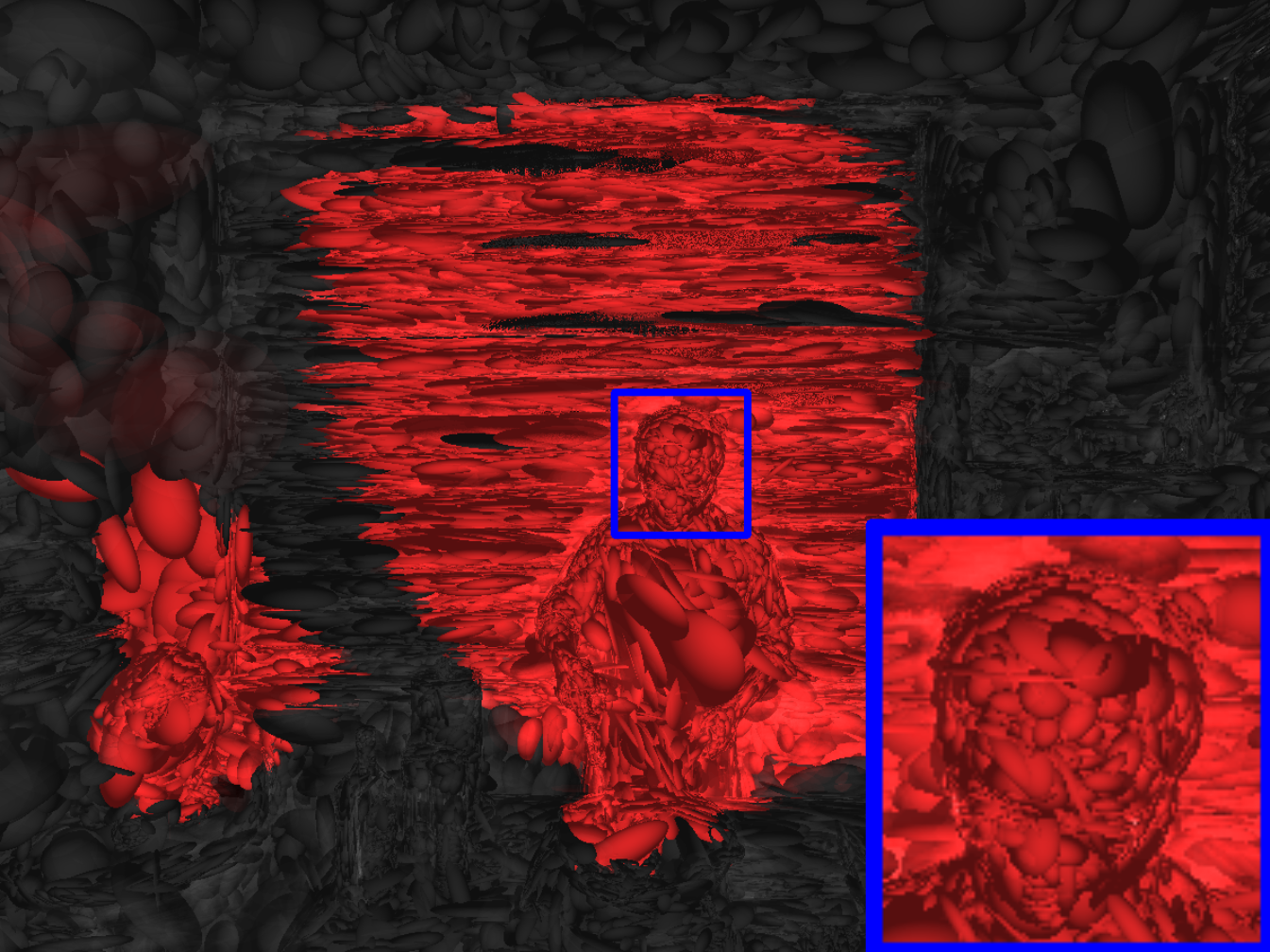}
                \caption{$G_m$}
            \end{minipage}
        \end{subfigure}
       \caption{Illustration of the clustering-based convex hull algorithm. (a) is the image from frame $t-1$. (b) and (c) are the rendered images without and with the convex hull algorithm, respectively. (d), (e), (f) shows the motion-related 3DGs. Specifically, (d) highlights the motion-related 3DGs via back-projection $G_o$, (e) is the 3DGs $G_i$ inside the convex structure, and (f) is the motio-related 3DGs $G_m$. It is clearly seen that ignoring $G_i$ would lead to the artifacts in (b).}
       \label{fig:Convex_Hull}
\end{figure}

As shown in \cref{fig:Convex_Hull} (d), GIM indexes the 3DGs $G_o$ whose locations are mostly distributed on the surface, and ignores those distributed inside the moving objects, causing the artifacts in the rendered images.
To address this, we introduce a clustering-based convex hull algorithm to find those motion-related 3DGs inside the moving objects. 
Specifically, we first perform positional clustering on $G_o$ with the clustering algorithm \cite{DBSCAN}:
\begin{equation}
      G_o^i\in L ,\ \text{if} \ \prod_{j}^{N} \delta(|(u_{G_o^i}-u_{G_o^j}|- \epsilon )=0,
      \label{eq:DBSCAN}
\end{equation}
where $i$ and $j$ represent the index of $G_o$, $L$ indicates the clustered \textit{Group}, $N$ is the number of 3DGs in group $L$, $u$ represents the position of 3DGs, and $\epsilon$ is the threshold, setting as $2$ empirically.
Then, we obtain the convex structure on each group of 3DGs using the algorithm \textit{Delaunay Triangulation} introduced in \cite{opencv}, and the 3DGs $G_i$ inside the convex structure are regarded as the motion-related 3DGs inside the moving object. All the motion-related 3DGs are represented as follows:
\begin{equation}
      G_m = \bigcup (G_o, G_i),
      \label{eq:DBSCAN}
\end{equation}
where $G_m$ is the motion-related 3DGs shown in \cref{fig:Convex_Hull} (f).

\begin{figure}[t!]  
	\centering
	\begin{subfigure}{\linewidth}
            \begin{minipage}[t]{0.31\linewidth}
                \centering
                \includegraphics[width=1\linewidth]{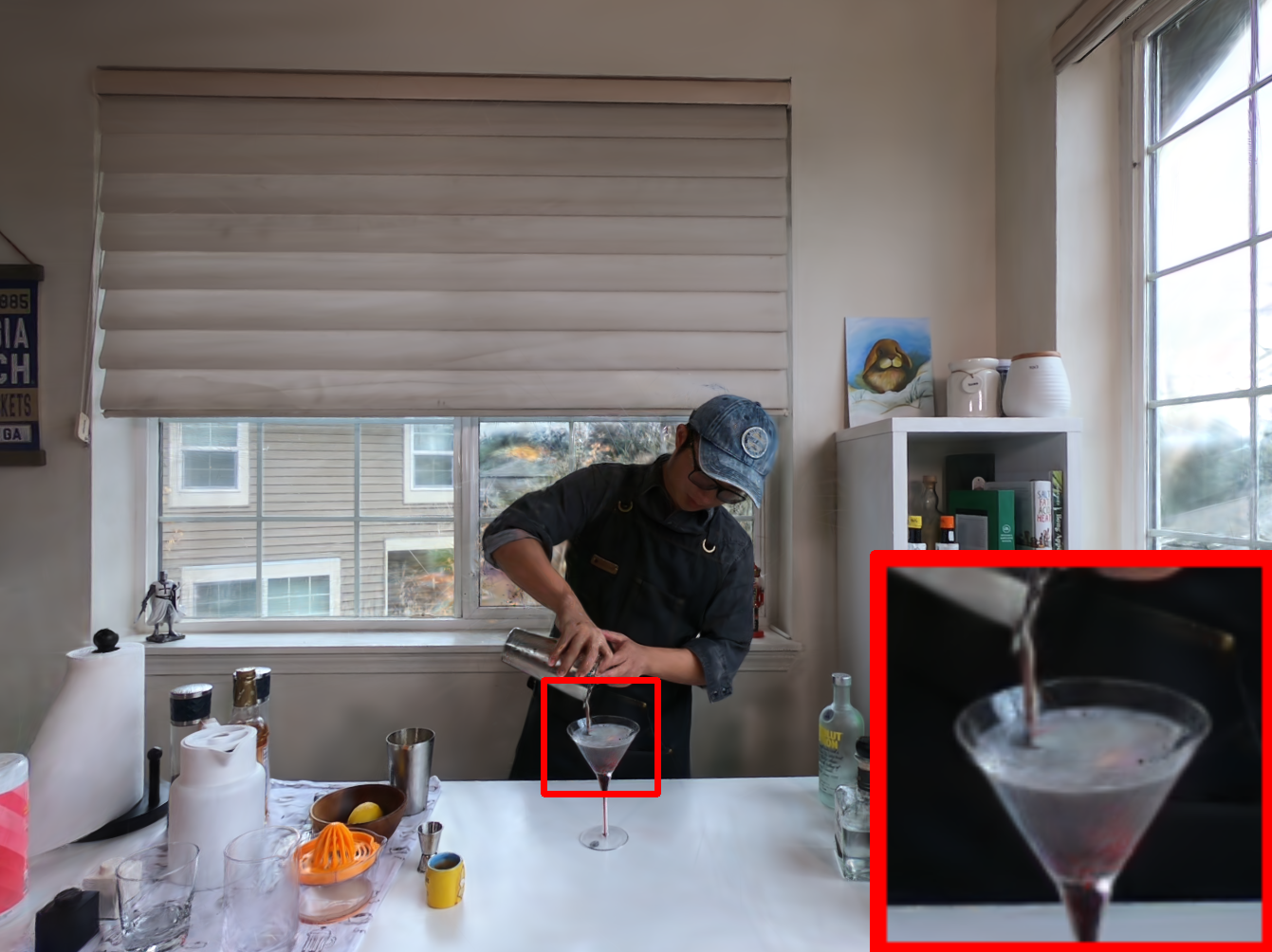}
            \end{minipage}
            \begin{minipage}[t]{0.31\linewidth}
                \centering
                \includegraphics[width=1\linewidth]{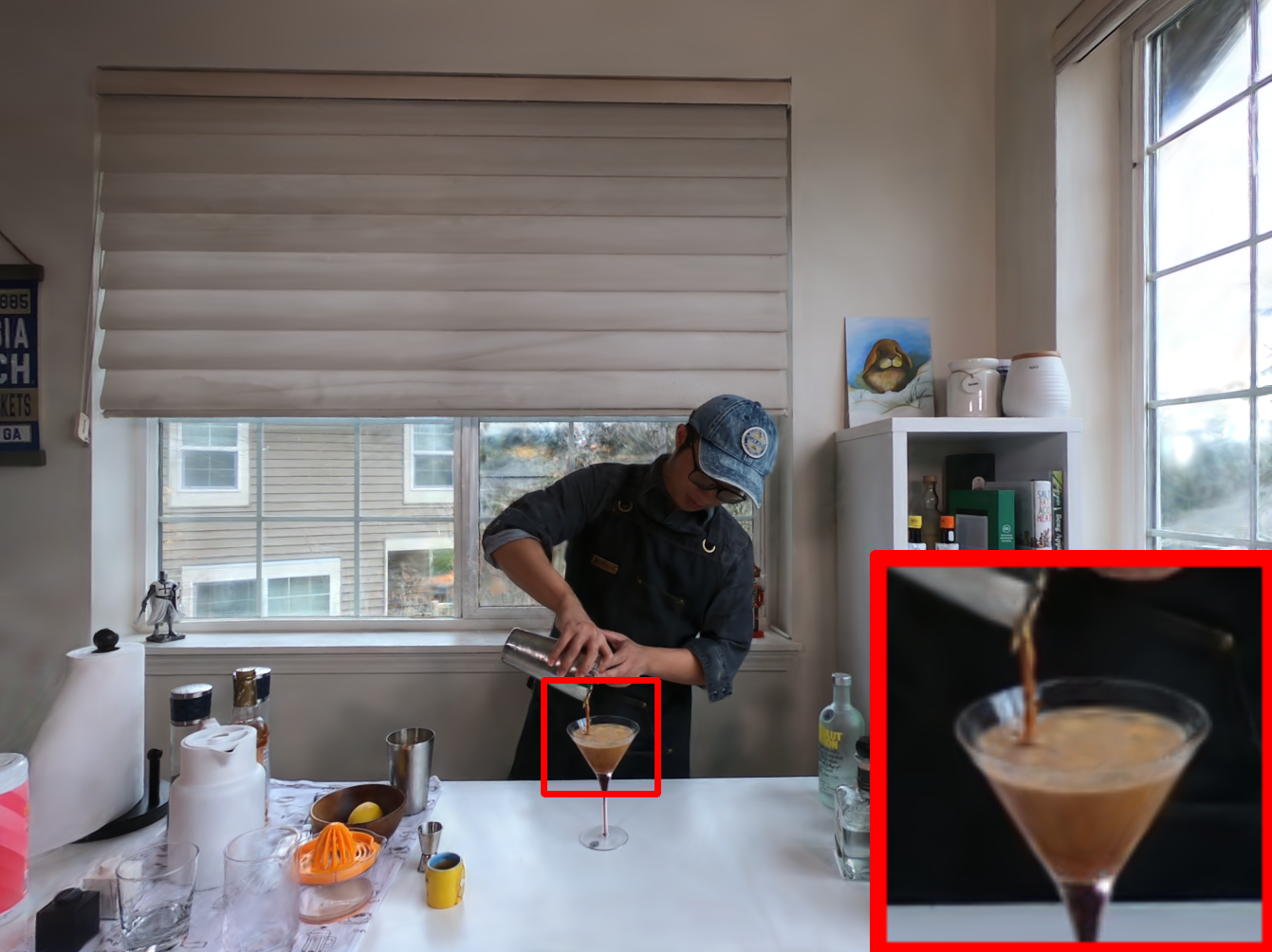}
            \end{minipage}
            \begin{minipage}[t]{0.31\linewidth}
                \centering
                \includegraphics[width=1\linewidth]{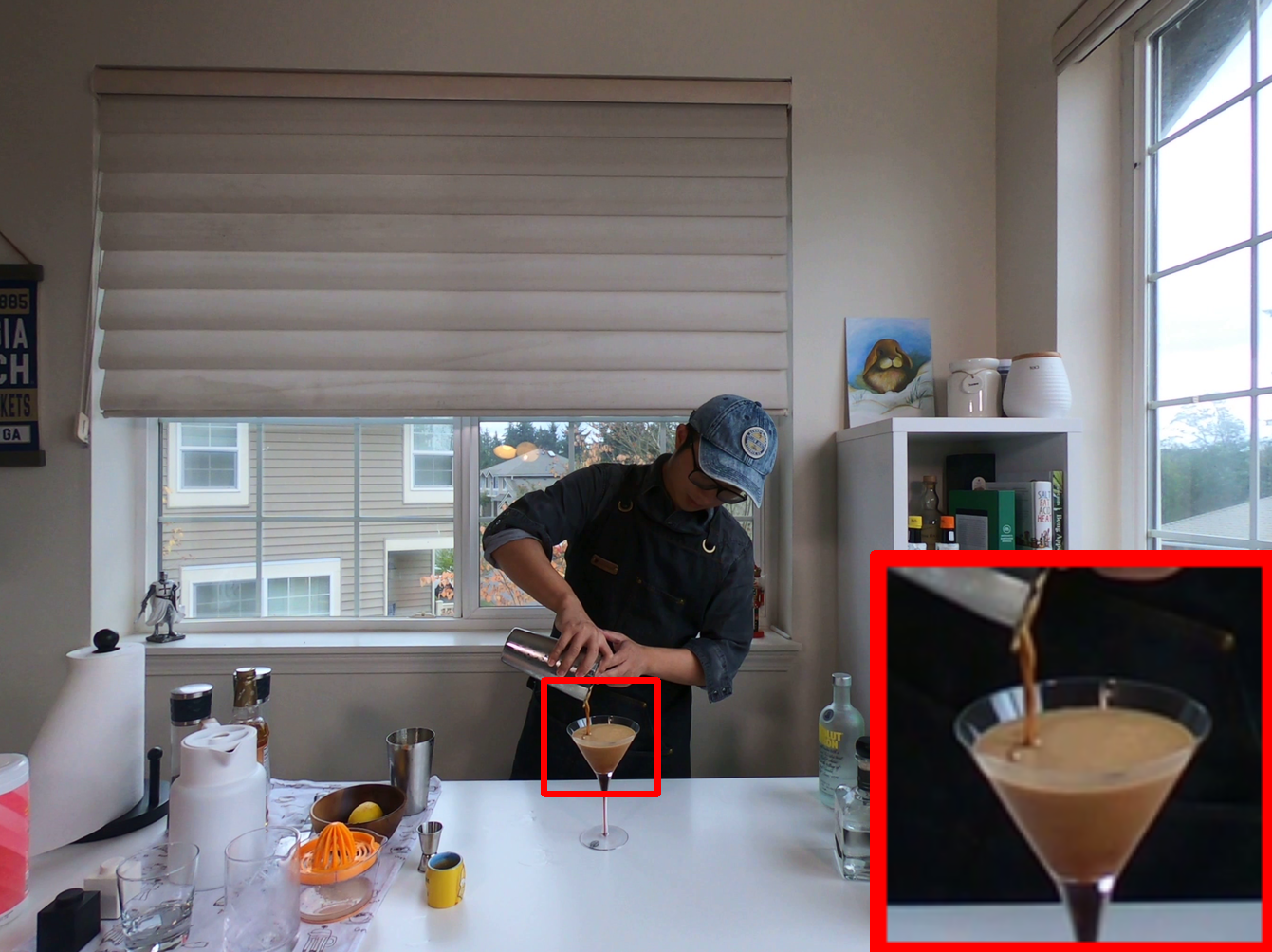}
            \end{minipage}
        \end{subfigure}

	\begin{subfigure}{\linewidth}
            \begin{minipage}[t]{0.31\linewidth}
                \centering
                \includegraphics[width=1\linewidth]{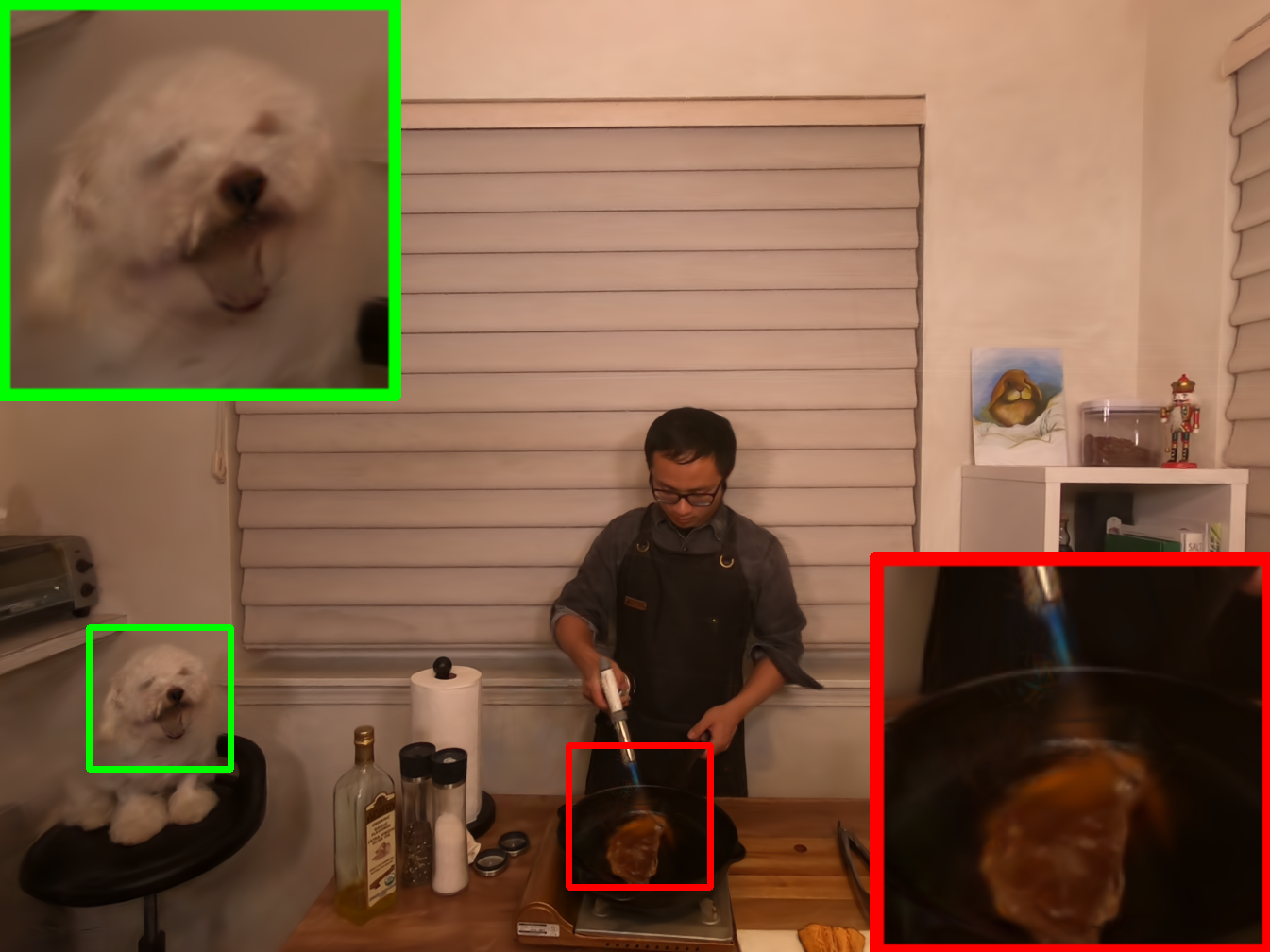}
                 \caption{Deformation}
            \end{minipage}
            \begin{minipage}[t]{0.31\linewidth}
                \centering
                \includegraphics[width=1\linewidth]{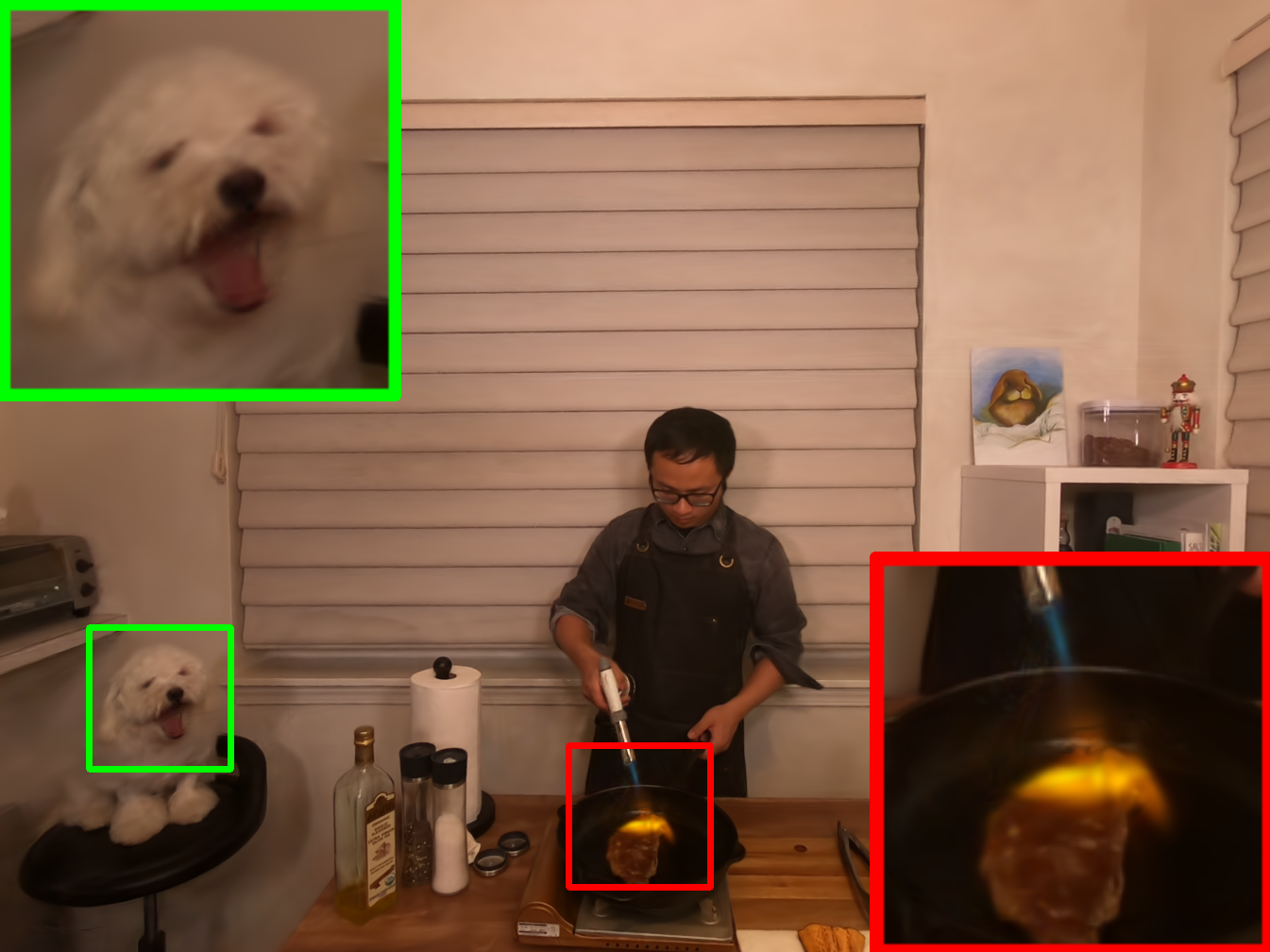}
                \caption{Optimization}
            \end{minipage}
            \begin{minipage}[t]{0.31\linewidth}
                \centering
                \includegraphics[width=1\linewidth]{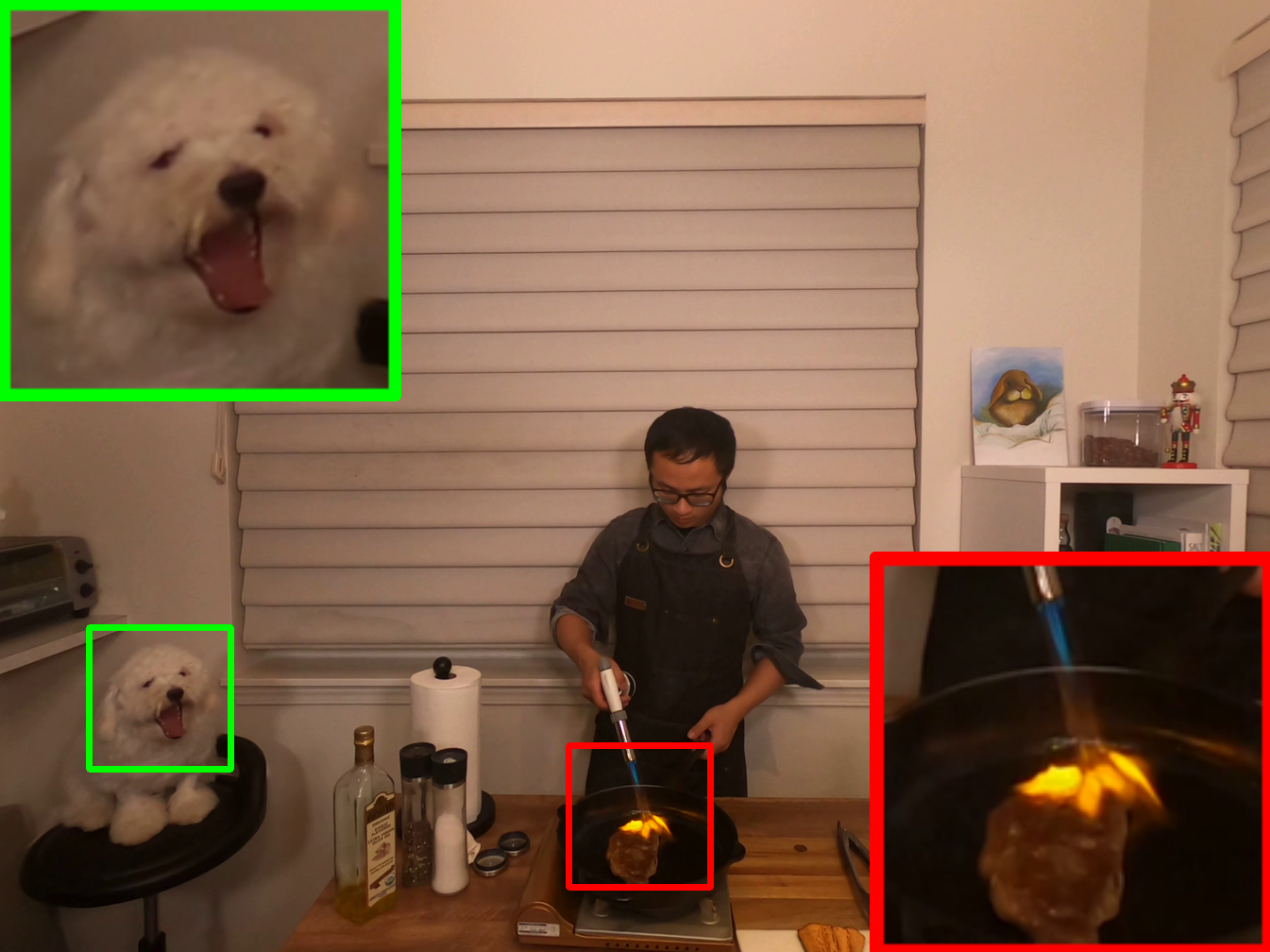}
                \caption{Ground Truth}
            \end{minipage}
        \end{subfigure}
       \caption{Illustration of Deformation and Optimization. }
       \label{fig:new_object}
\end{figure}



\subsection{3DGs Deformation and Optimization} \label{method_deformation_optimization} 
In order to model the dynamic scenes, the motion-related 3DGs $G_m$ need the deformation operation.
Following the previous methods \cite{3dgstream,dynamic3dgs}, we apply rigid transformations on $G_m$ for computational and storage efficiency, as follows: 
\begin{equation}
\begin{aligned}
      \Delta u_t, \Delta q_t &= \mathcal{F}_d  (u_{G_m}), \\
      u_t, q_t = u_{t-1} + & \Delta u_t, q_{t-1} \times \Delta q_t,
      \label{eq:deform}
\end{aligned}
\end{equation}
where $u_{G_m}$ represents the position of $G_m$ at timestep $t-1$, $F_d$ represents the mapping function for deformation, and $\Delta u_t$ and $\Delta q_t$ denote the translational and rotational offsets relative to the previous frame, respectively. The mapping function $\mathcal{F}_d$  is implemented using hash grids, which is optimized with the objective $\mathcal{L}_{color}$ in equation \eqref{eq:3dgs_loss}. 

As seen in \cref{fig:new_object}, the deformation in \cref{eq:deform} can successfully capture the dynamic, \eg, hand movement, but fail to model the emerging objects correctly, \ie, the liquid in the goblet shows different color from that in the real scene.
The reason is that the spherical harmonic coefficient $sh$ in $G_m$ is initialized based on the previous frame, and they are incapable of predicting the precise color for the emerging object.
To solve this, we optimize the spherical harmonic coefficient for modeling the emerging objects. 

\begin{figure*}[ht!]
	\centering

	\begin{subfigure}{\linewidth}
            \rotatebox[origin=c]{90}{\footnotesize{Discussion}\hspace{-2.6cm}}
            \begin{minipage}[t]{0.193\linewidth}
                \centering
                \includegraphics[width=1\linewidth]{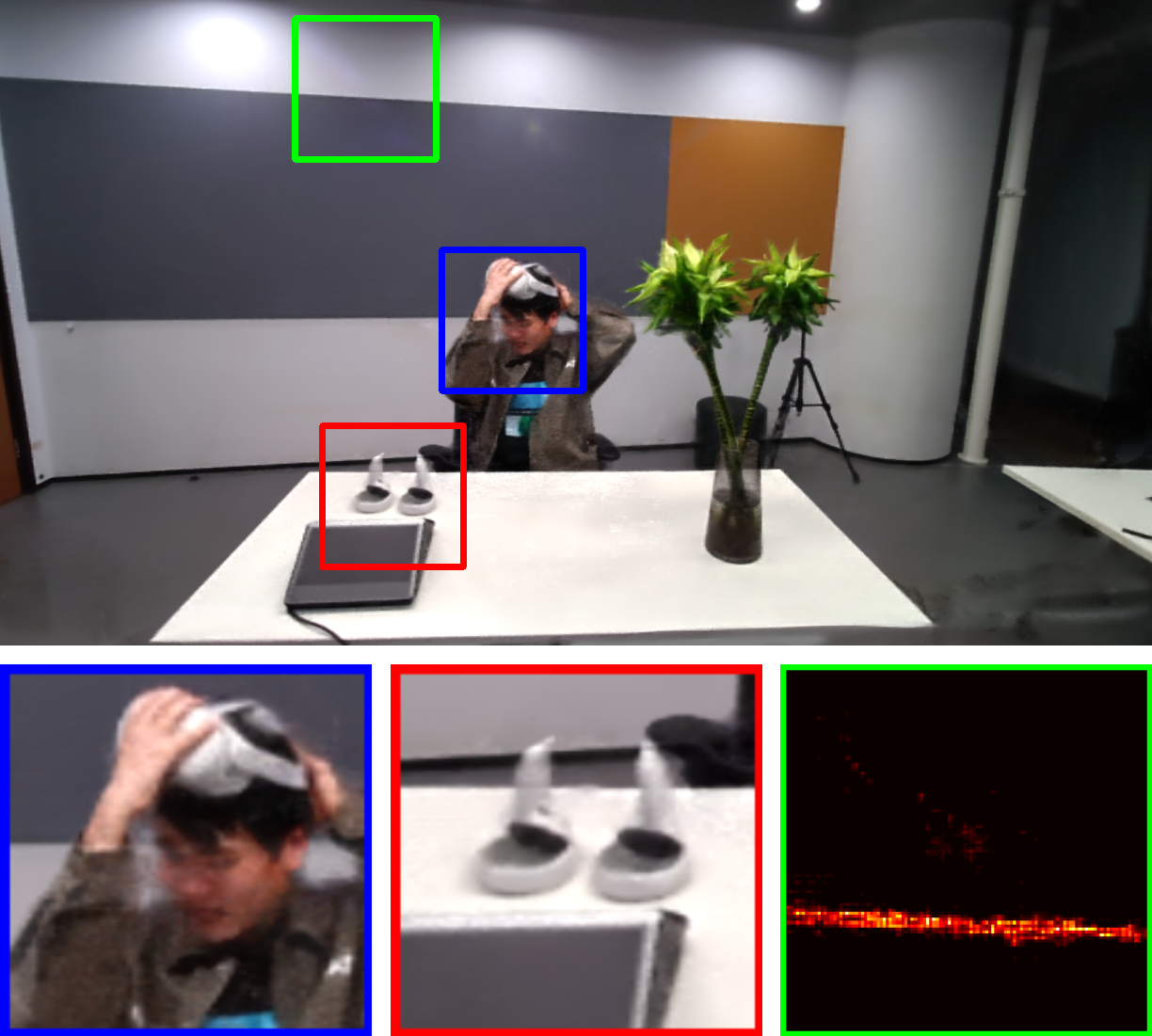}
            \end{minipage}
            \begin{minipage}[t]{0.193\linewidth}
                \centering
                \includegraphics[width=1\linewidth]{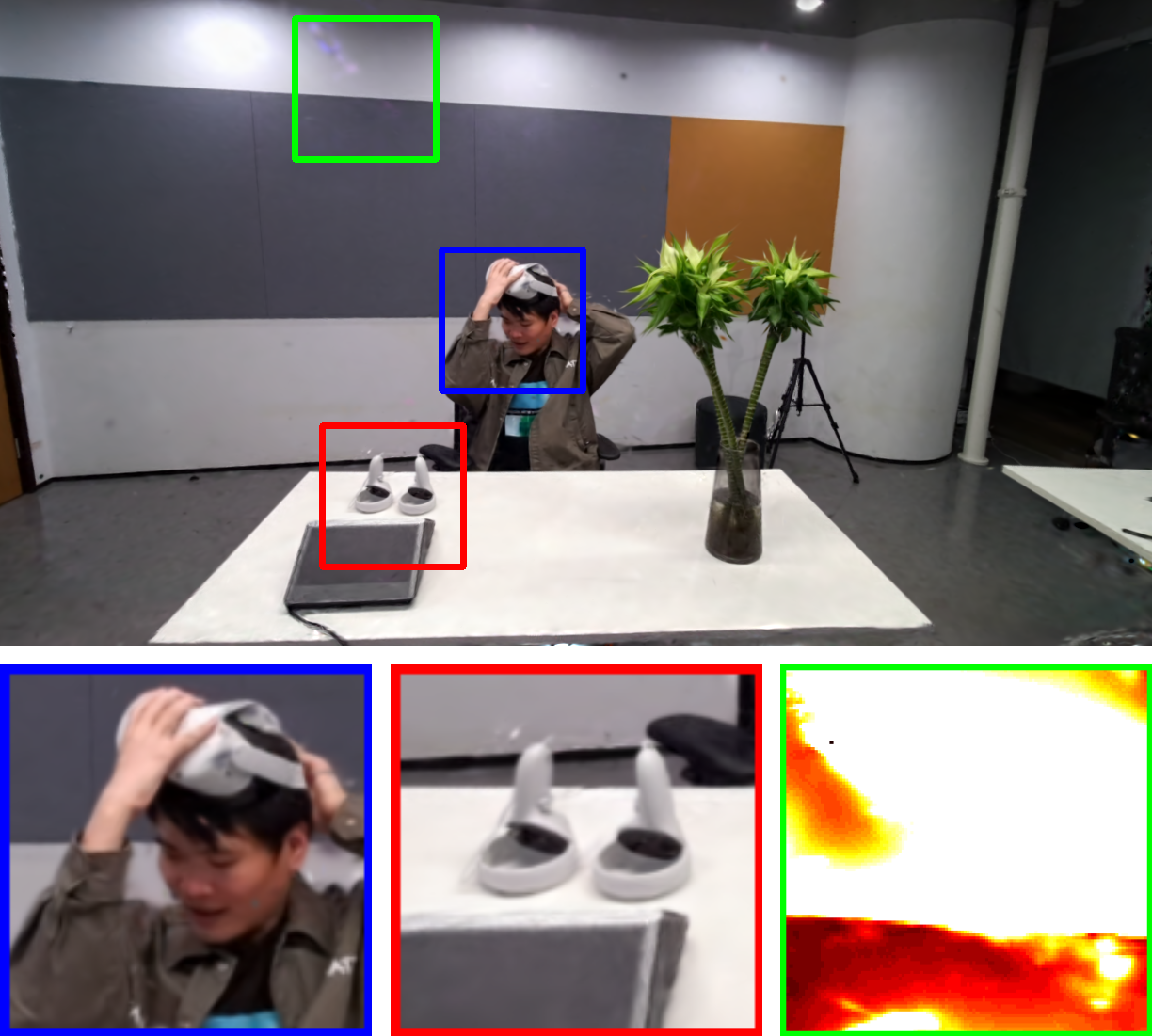}
            \end{minipage}
            \begin{minipage}[t]{0.193\linewidth}
                \centering
                \includegraphics[width=1\linewidth]{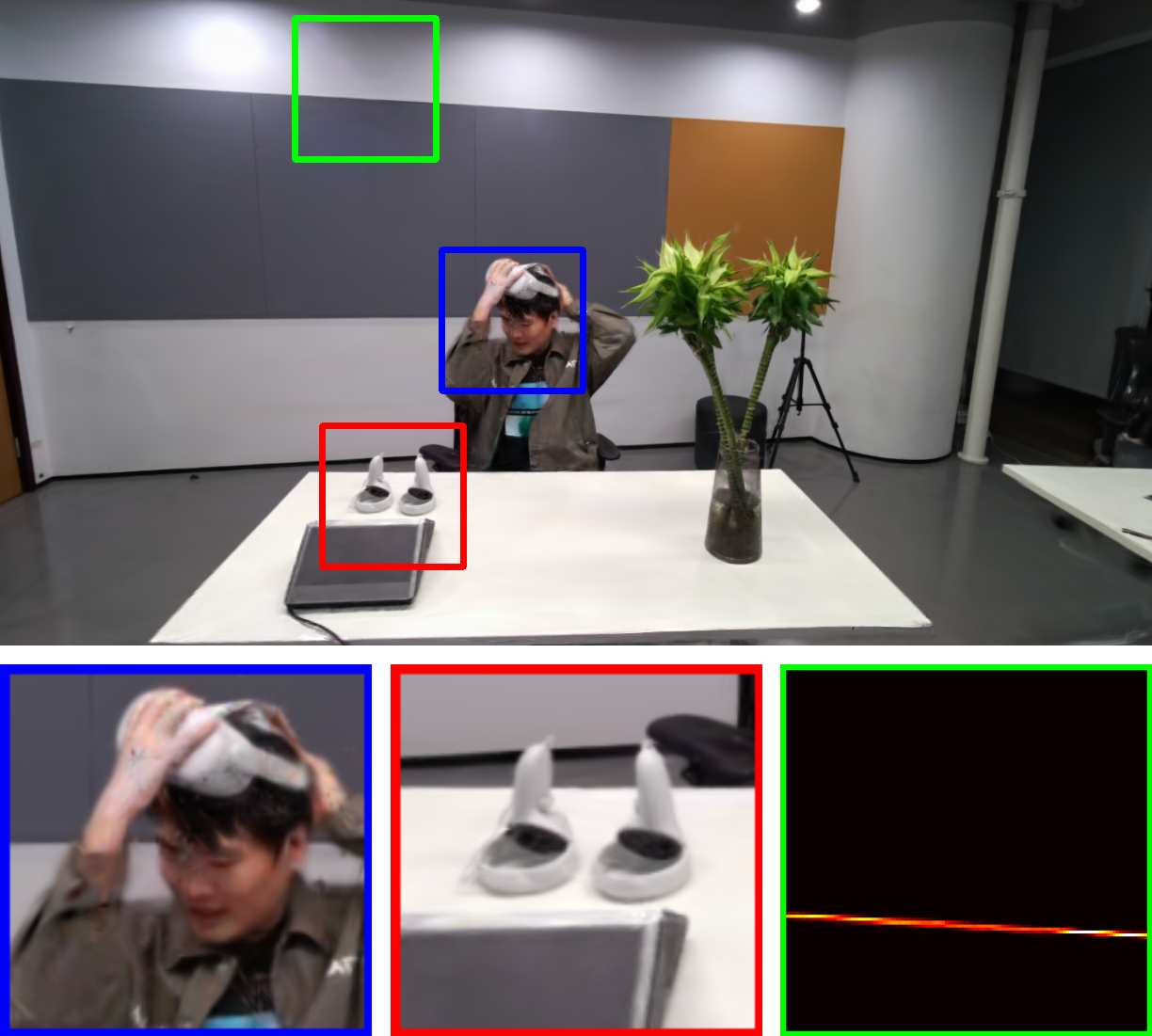}
            \end{minipage}
            \begin{minipage}[t]{0.193\linewidth}
                \centering
                \includegraphics[width=1\linewidth]{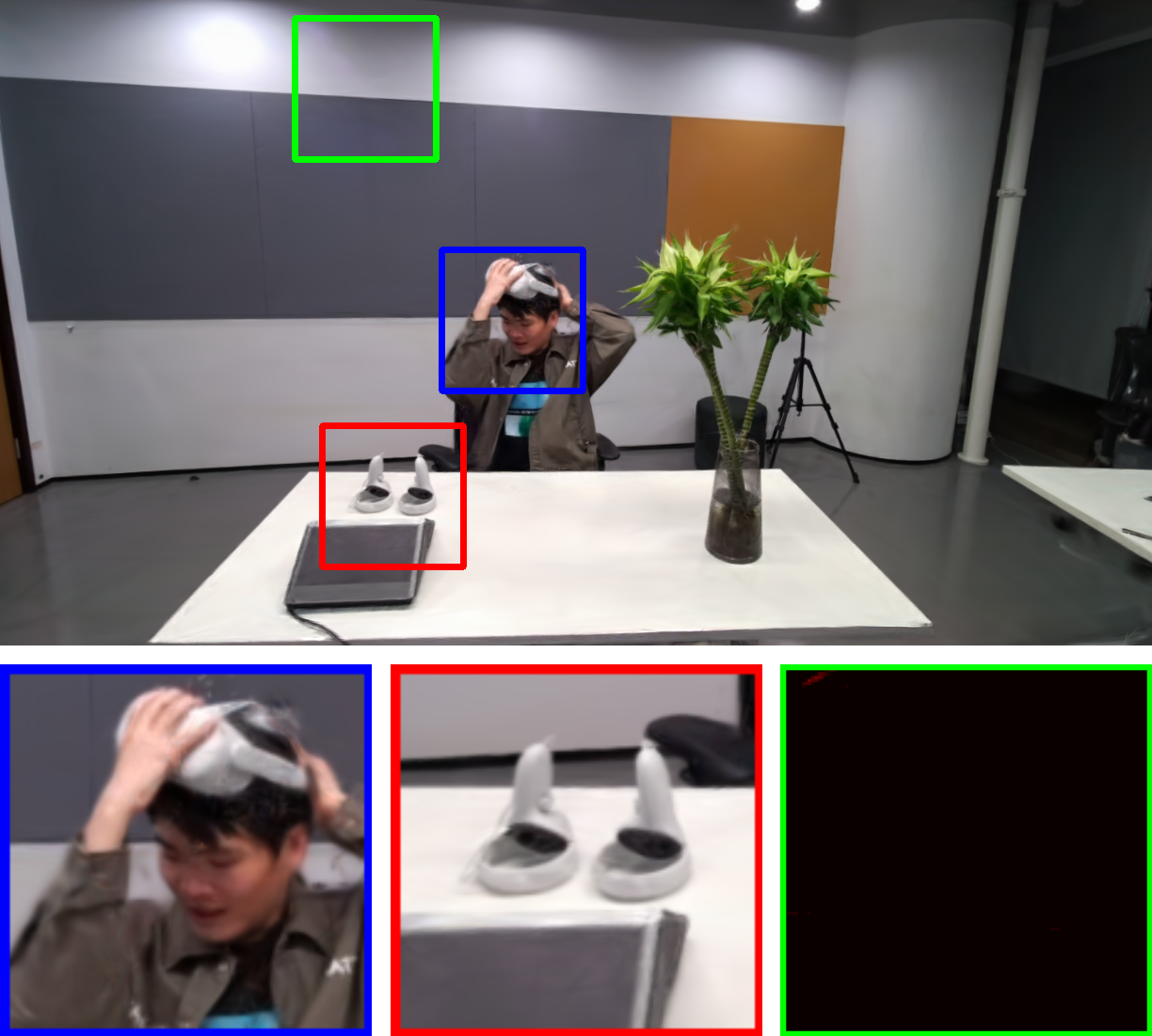}
            \end{minipage}
            \begin{minipage}[t]{0.193\linewidth}
                \centering
                \includegraphics[width=1\linewidth]{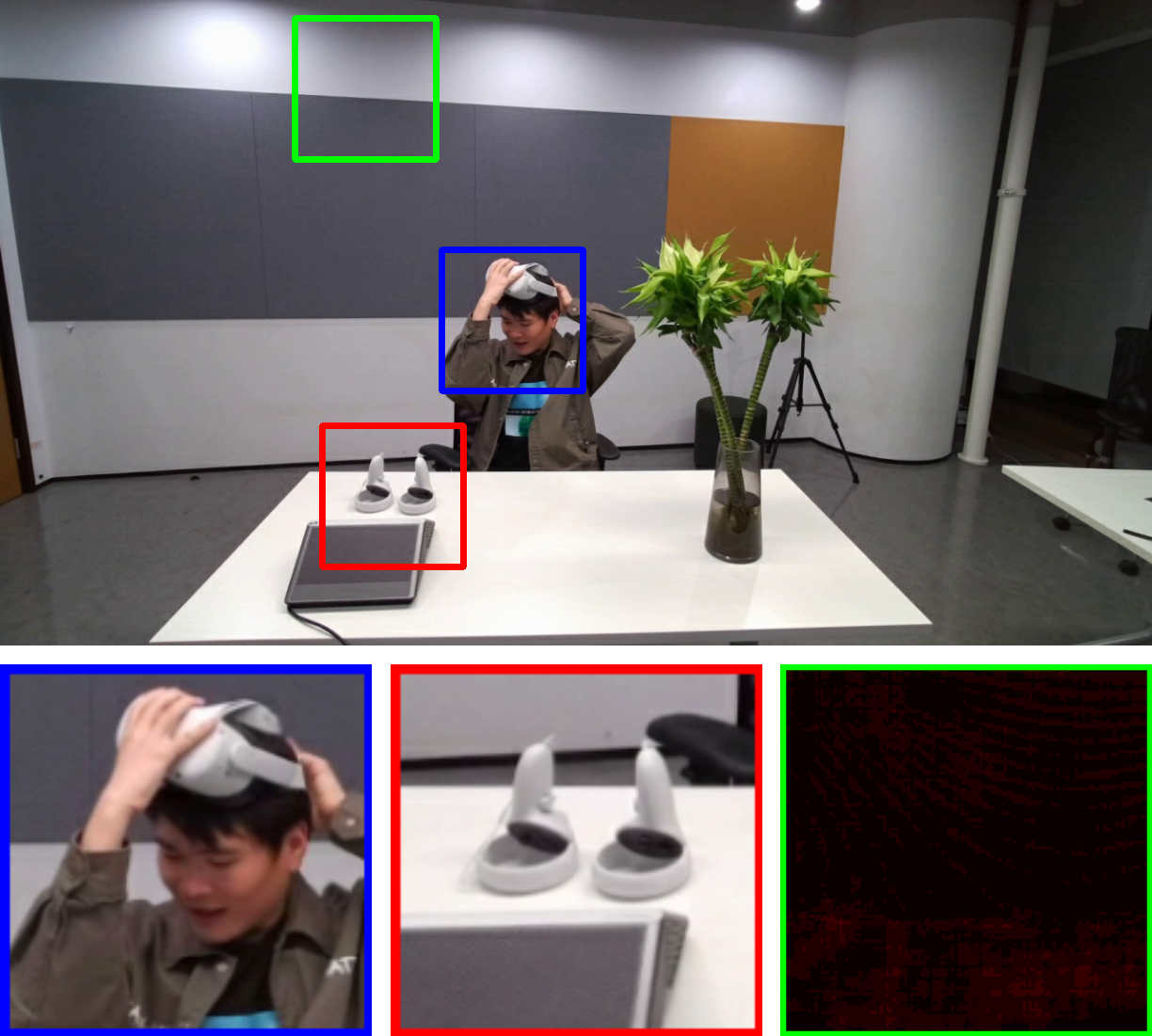}
            \end{minipage}
        \end{subfigure}

	\begin{subfigure}{\linewidth}
            \rotatebox[origin=c]{90}{\footnotesize{Coffee Martini}\hspace{-3.6cm}}
            \begin{minipage}[t]{0.193\linewidth}
                \centering
                \includegraphics[width=1\linewidth]{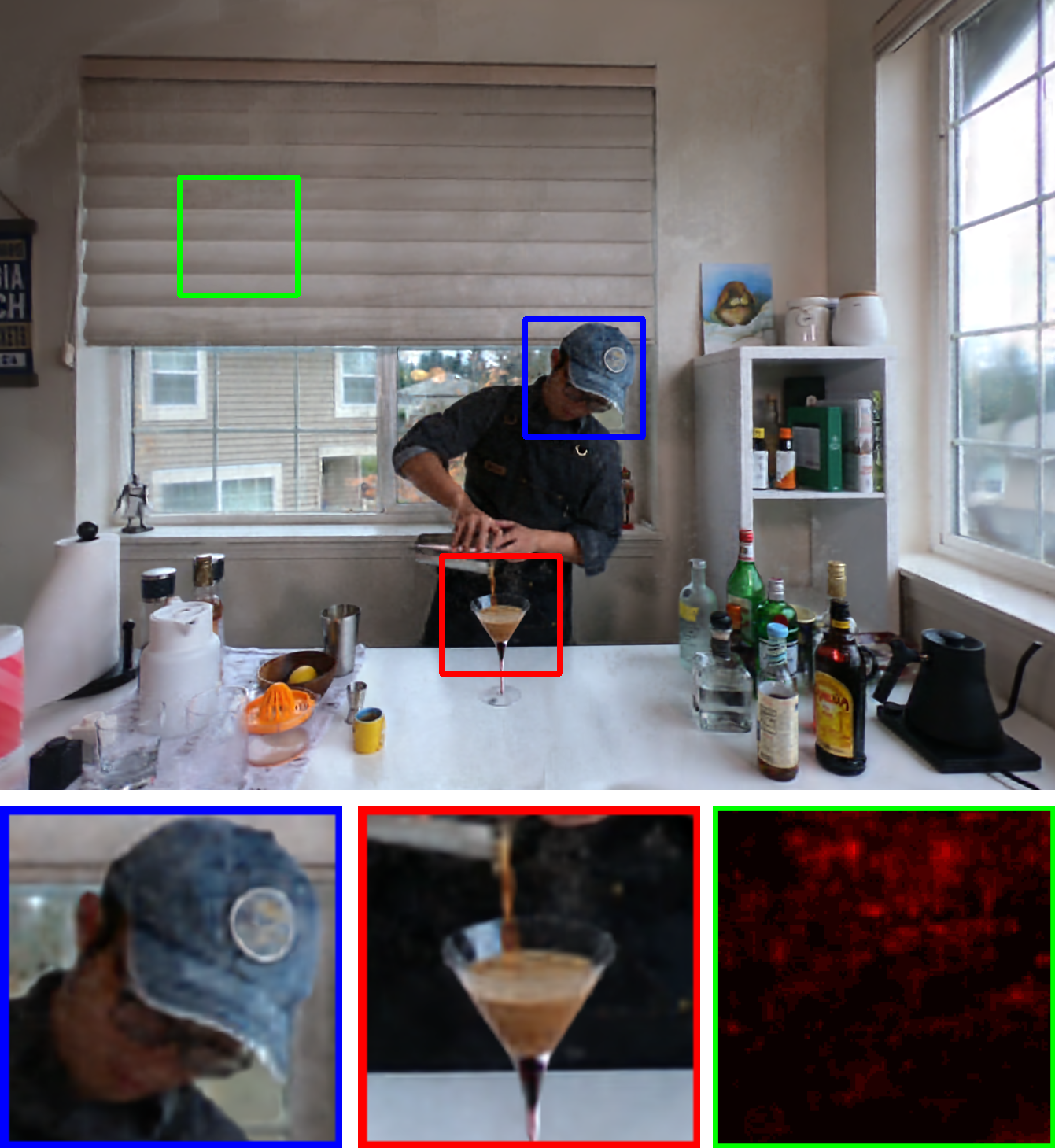}
            \end{minipage}
            \begin{minipage}[t]{0.193\linewidth}
                \centering
                \includegraphics[width=1\linewidth]{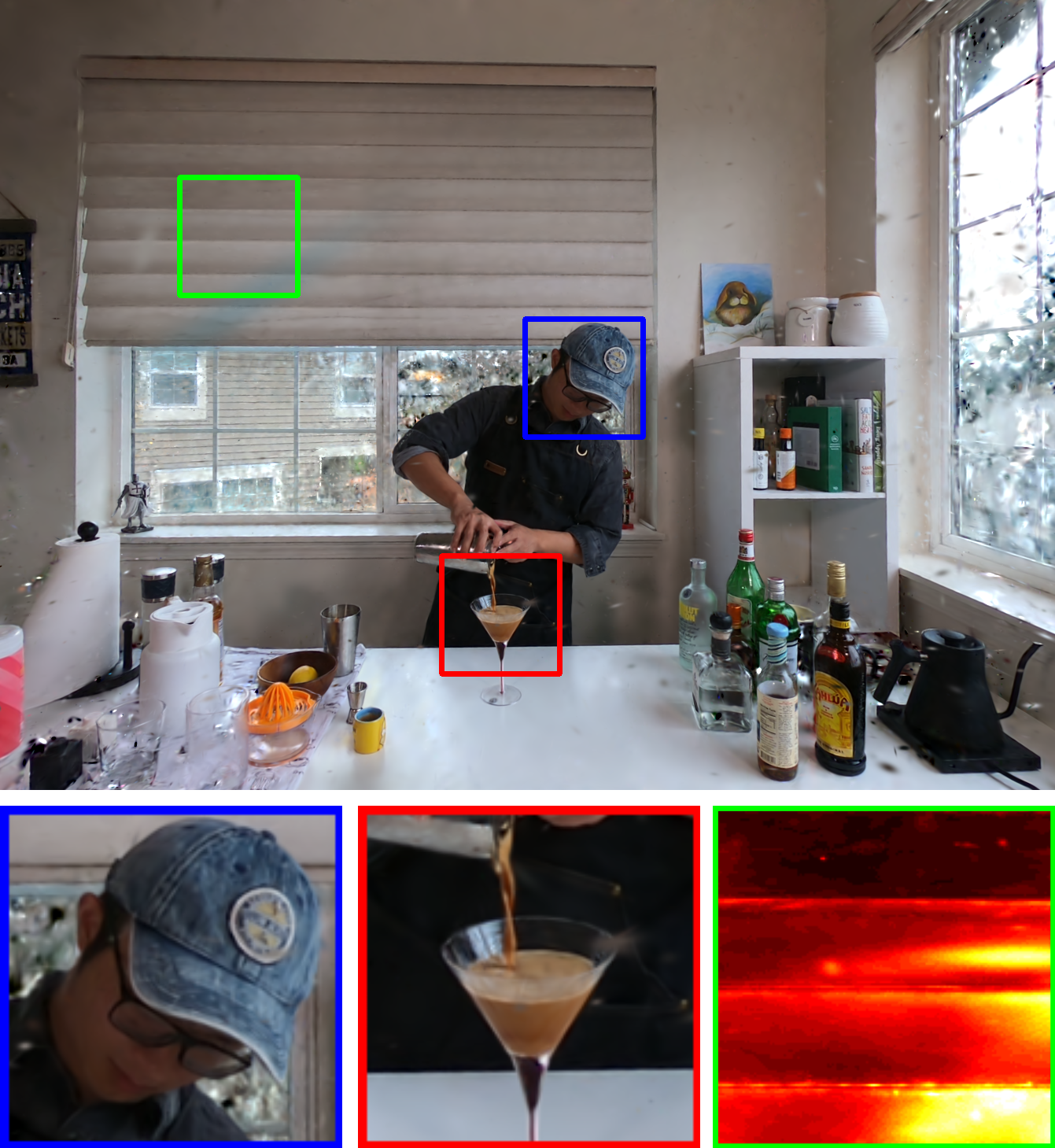}
            \end{minipage}
            \begin{minipage}[t]{0.193\linewidth}
                \centering
                \includegraphics[width=1\linewidth]{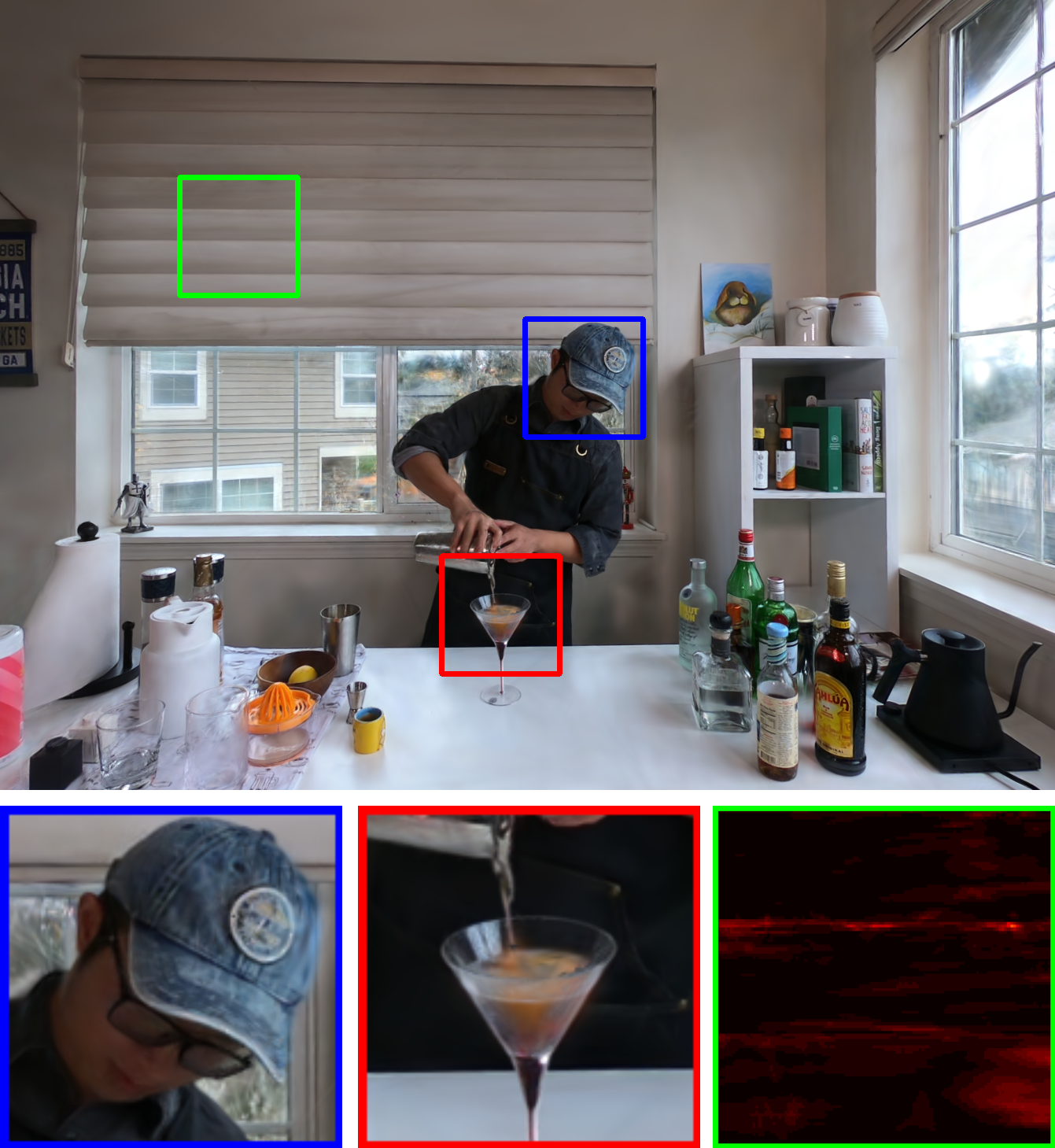}
            \end{minipage}
            \begin{minipage}[t]{0.193\linewidth}
                \centering
                \includegraphics[width=1\linewidth]{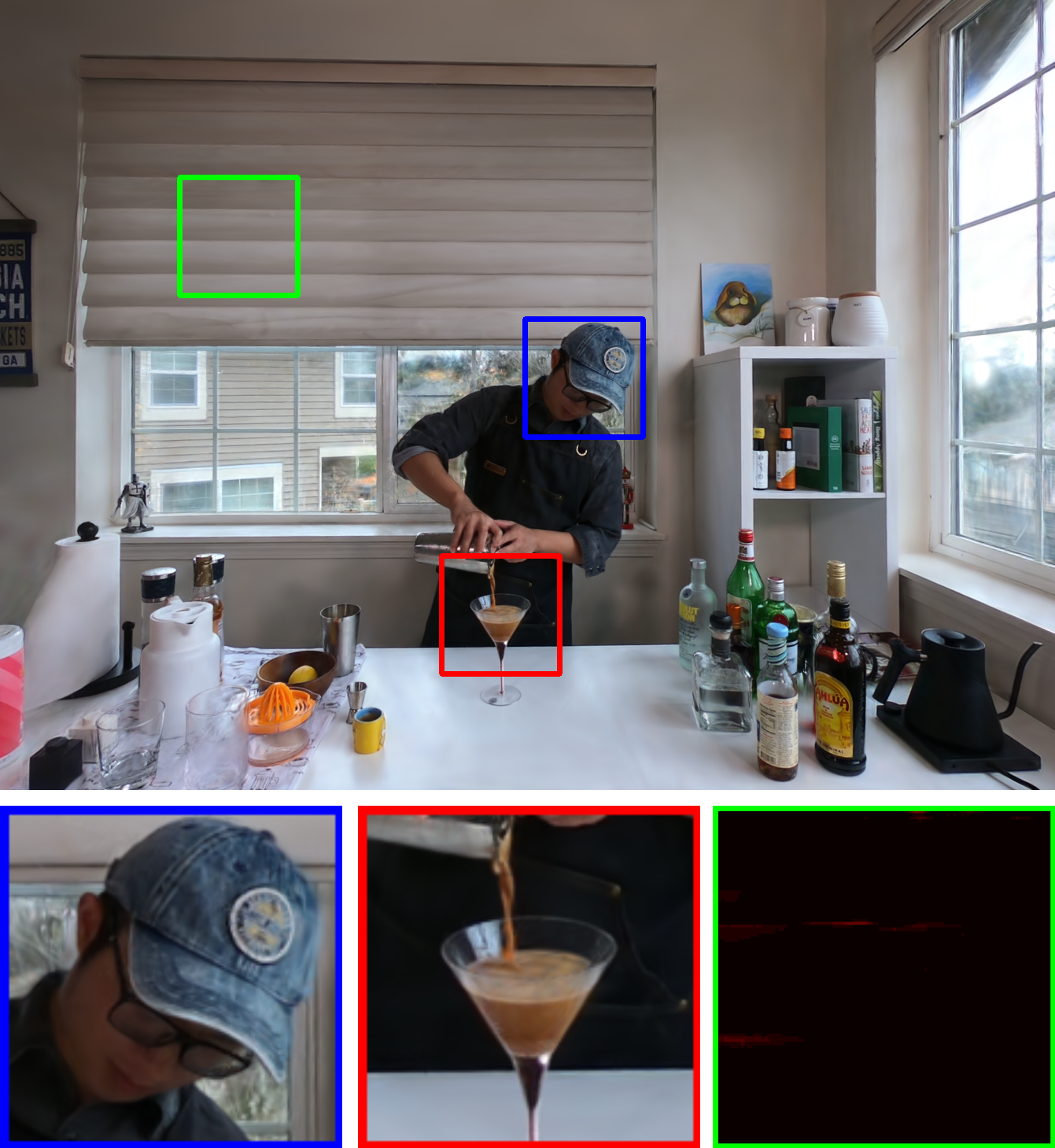}
            \end{minipage}
            \begin{minipage}[t]{0.193\linewidth}
                \centering
                \includegraphics[width=1\linewidth]{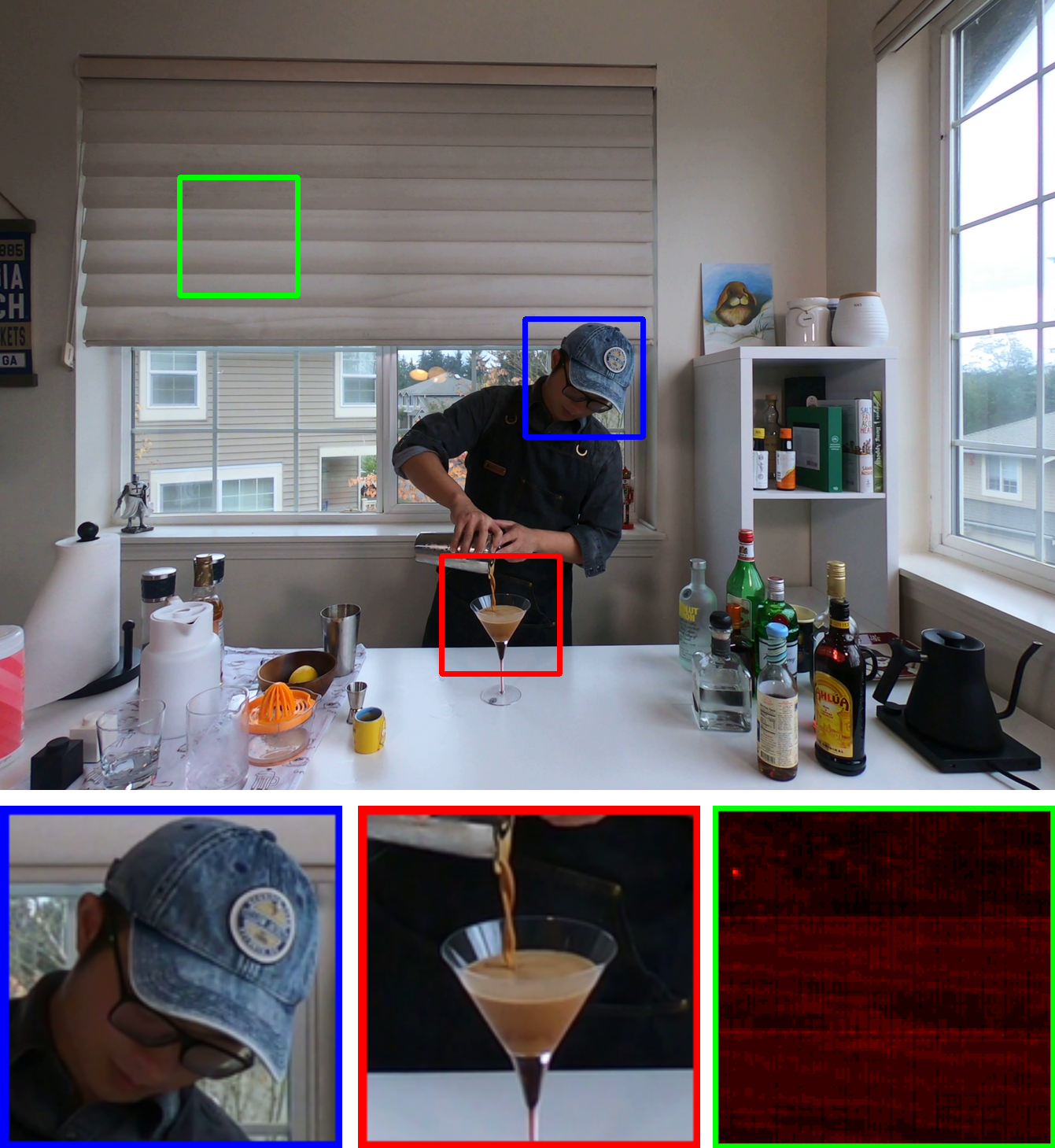}
            \end{minipage}
        \end{subfigure}

	\begin{subfigure}{\linewidth}
            \rotatebox[origin=c]{90}{\footnotesize{Cook SPinach}\hspace{-3.4cm}}
            \begin{minipage}[t]{0.193\linewidth}
                \centering
                \includegraphics[width=1\linewidth]{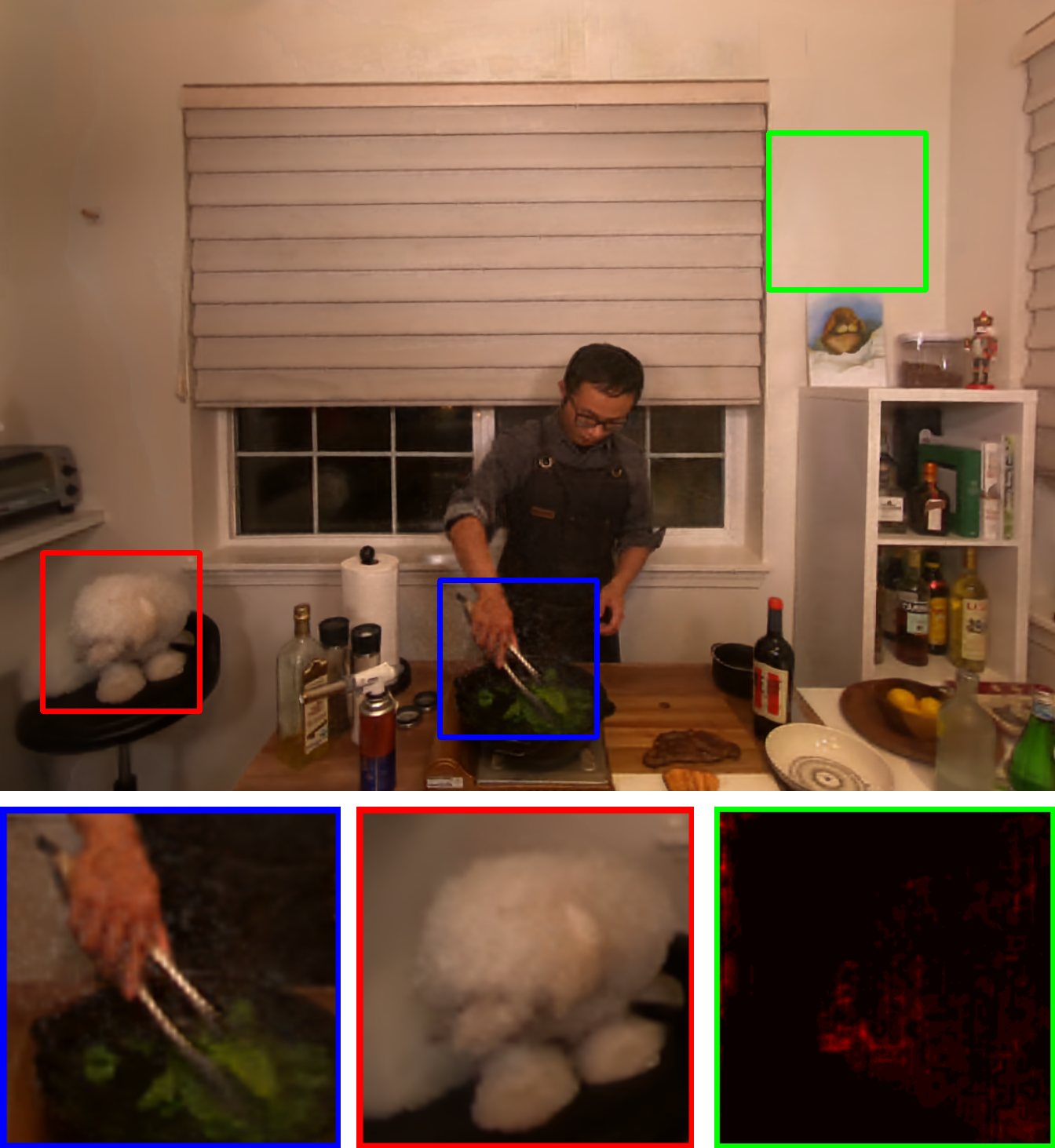}
                \caption{StreamRF}
            \end{minipage}
            \begin{minipage}[t]{0.193\linewidth}
                \centering
                \includegraphics[width=1\linewidth]{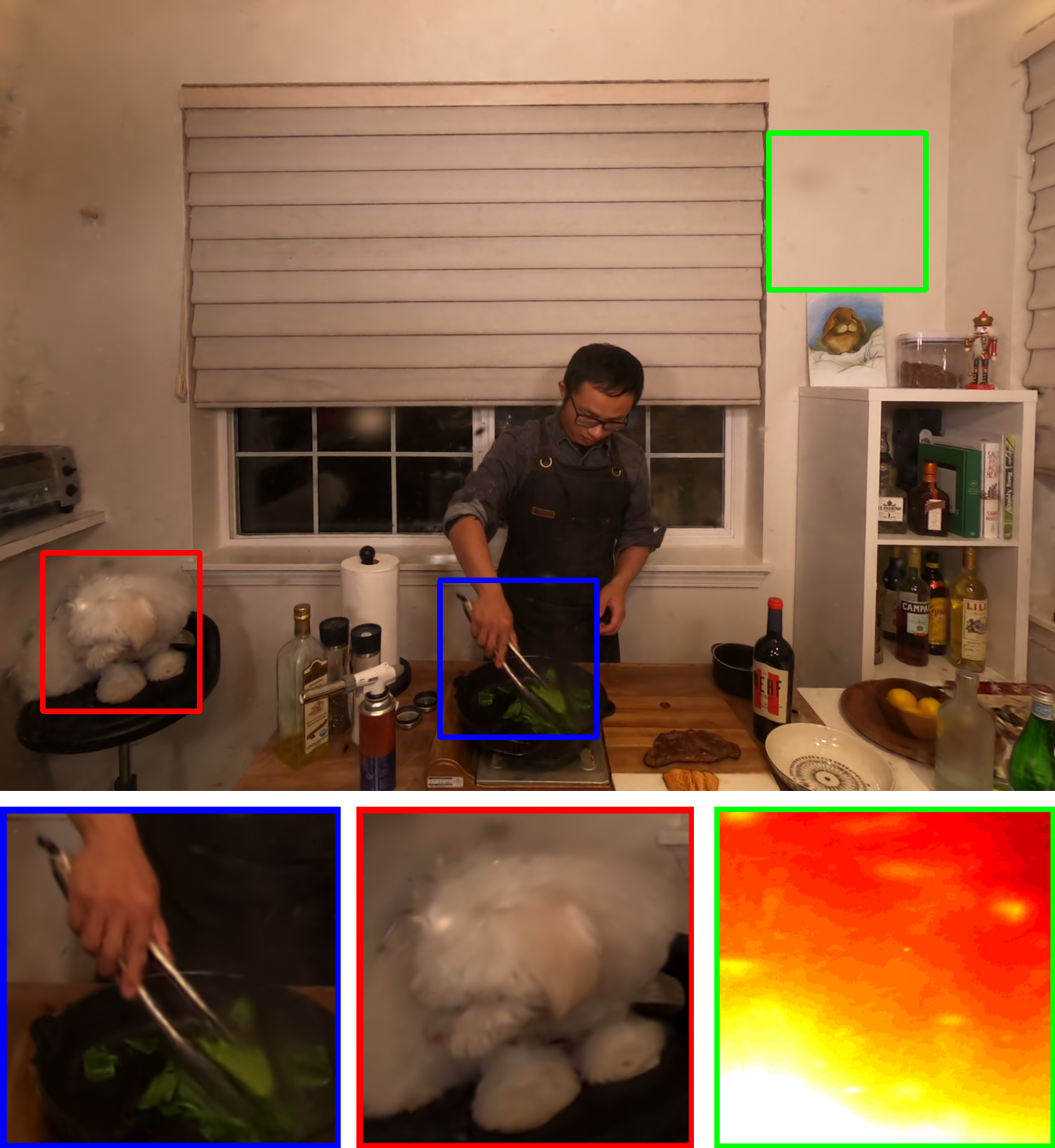}
                \caption{Dynamic-3DGS}
            \end{minipage}
            \begin{minipage}[t]{0.193\linewidth}
                \centering
                \includegraphics[width=1\linewidth]{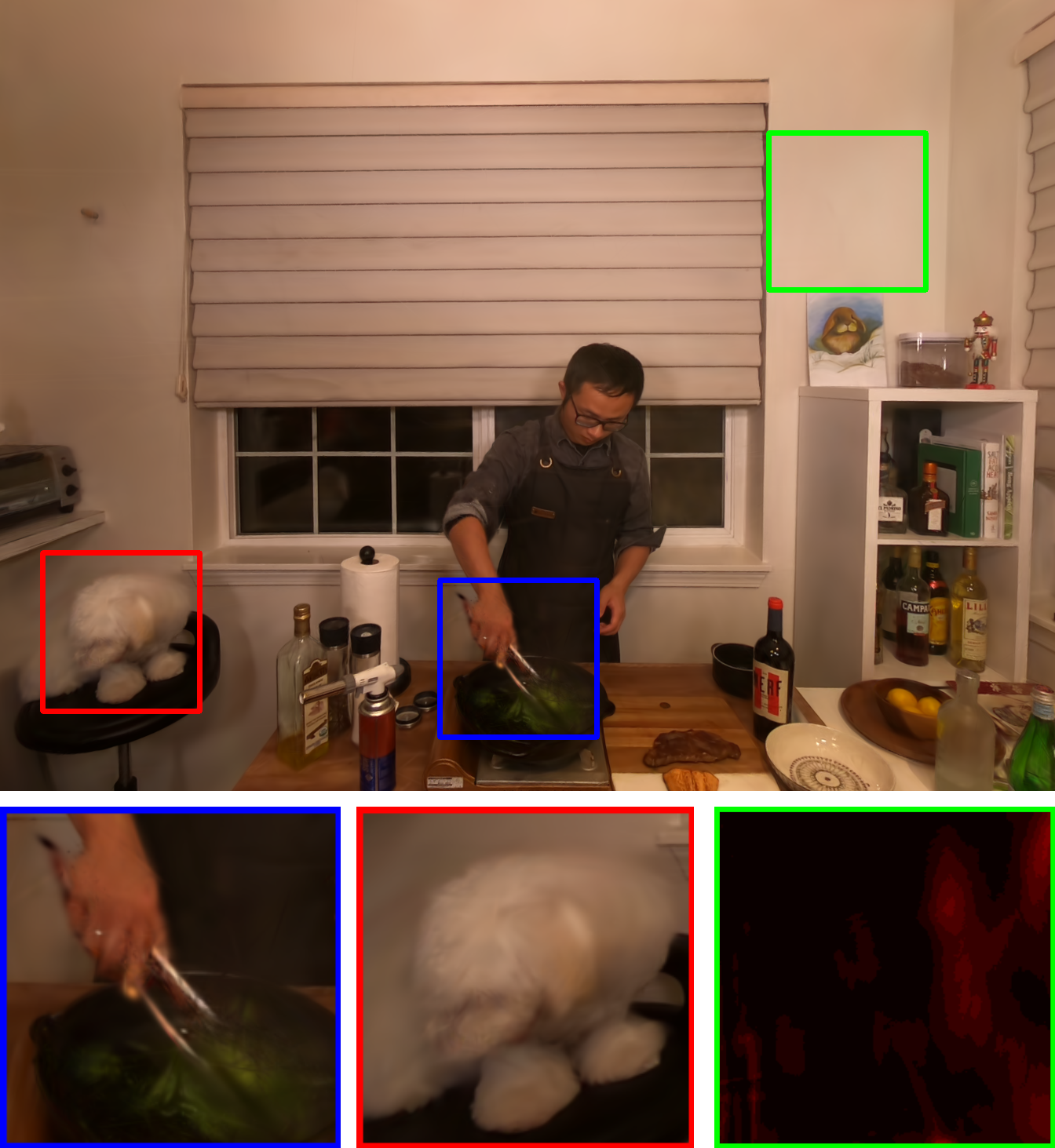}
                \caption{3DGStream}
            \end{minipage}
            \begin{minipage}[t]{0.193\linewidth}
                \centering
                \includegraphics[width=1\linewidth]{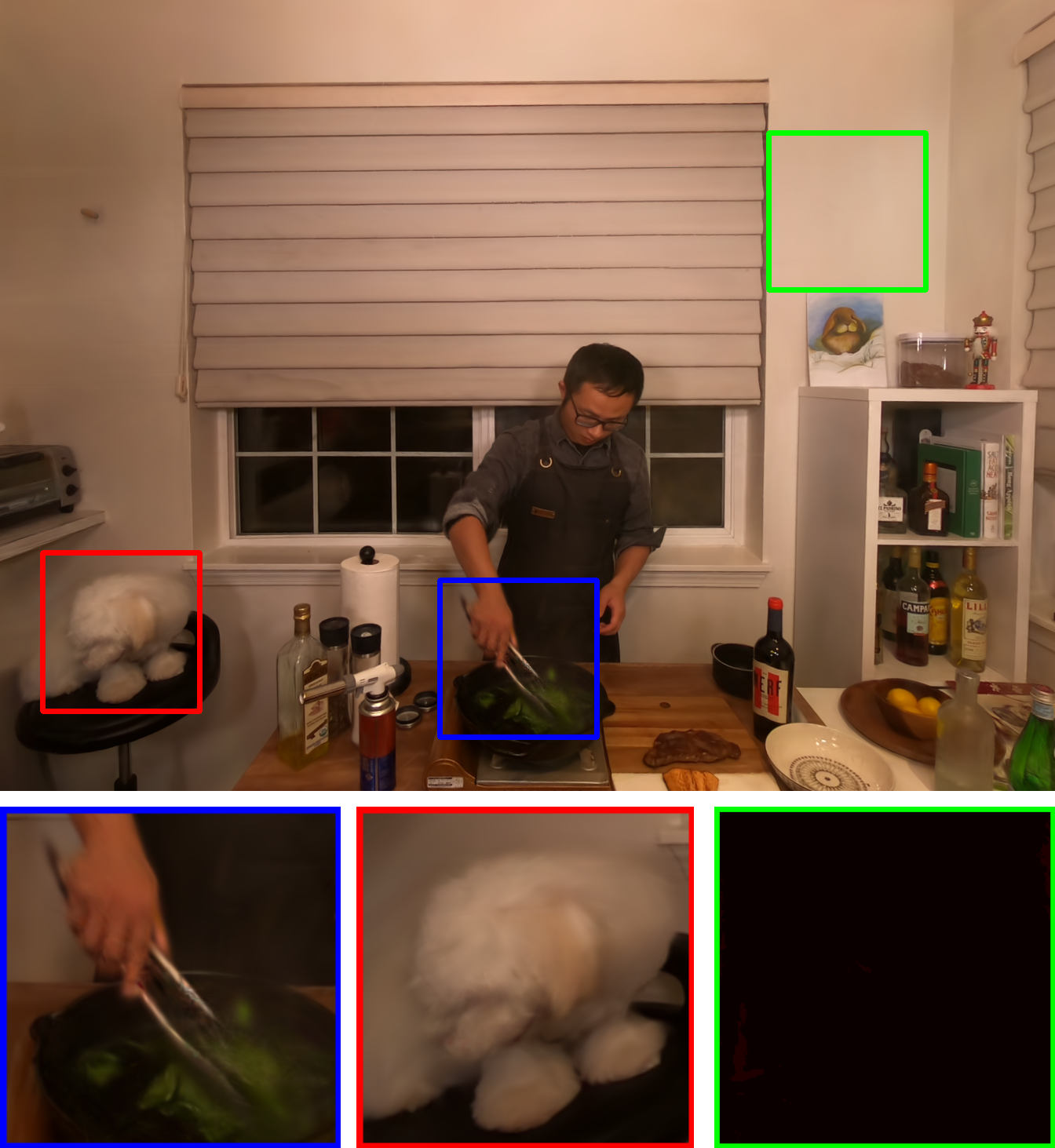}
                \caption{Ours}
            \end{minipage}
            \begin{minipage}[t]{0.193\linewidth}
                \centering
                \includegraphics[width=1\linewidth]{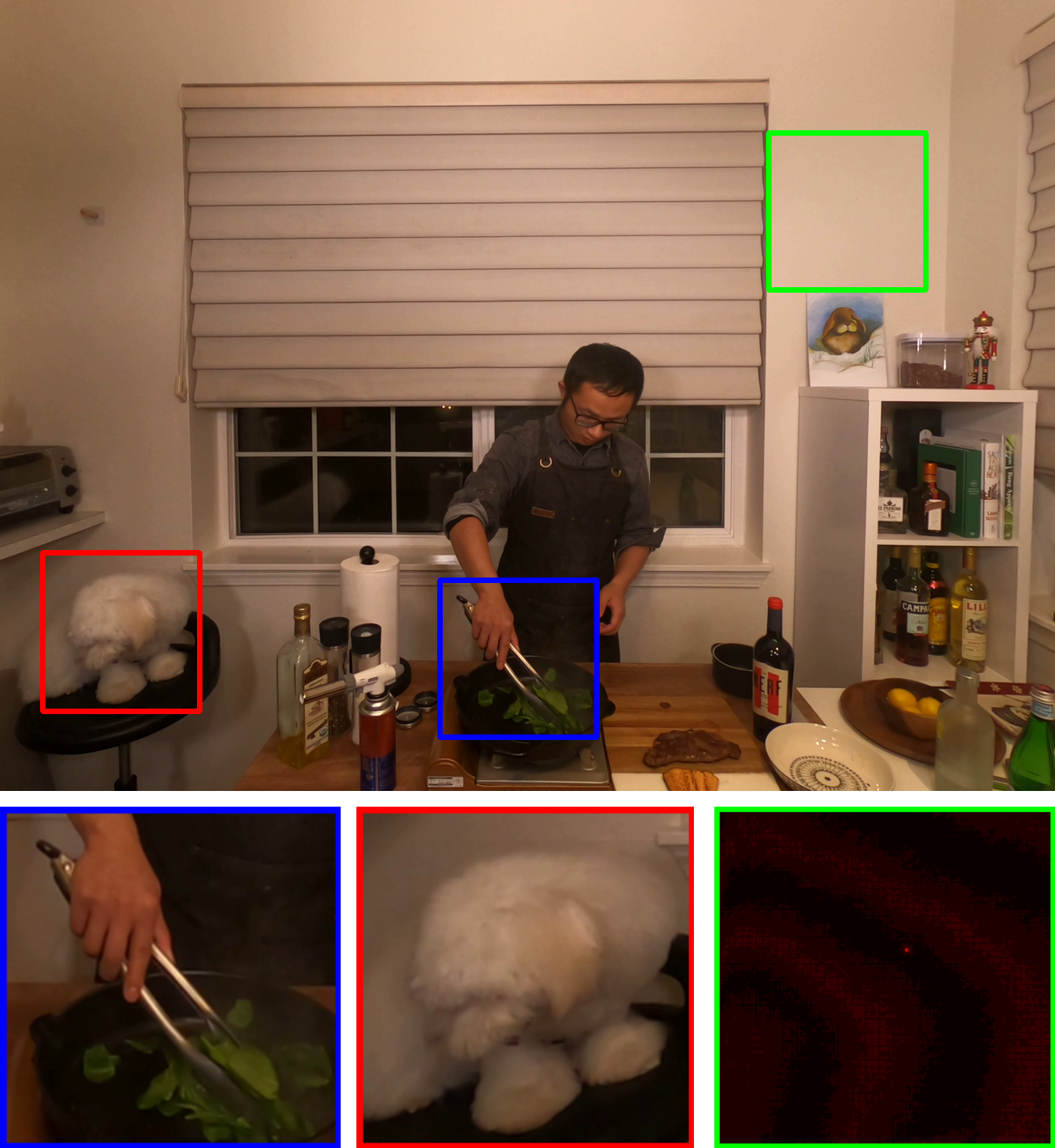}
                \caption{Ground Truth}
            \end{minipage}
        \end{subfigure}
	\caption{Qualitative Results on Real World Datasets MeetRoom (\textit{Disscussion}) and N3DV (\textit{Coffee Martini} and \textit{Cook Spinach}). The \textcolor{blue}{blue} and \textcolor{red} {red} boxes show the quality of the novel synthesis, and \textcolor{green} {green} boxes represent the heatmap of the temporal variance in the static region, where the blacker the color, the smaller the variance.
    }
	\label{fig:compare_figure}
\end{figure*}

\begin{table*}[!t]
\caption{Quantitative comparison with online methods on the N3DV and MeetRoom datasets. The PSNR, storage memory, training time and $E_{warp}$ are averaged over 300 frames for each scene. 
PSNR($\cdot$ / $\cdot$) means the PSNR calculated using the entire image and the dynamic region, respectively. The detail abouts PSNR calculation can be found in the supplementary material.
\colorbox{Magenta!30}{Pink} and \colorbox{Melon!40}{Brown} indicate the best and the second, respectively. * means re-implementing the experiments with identical initiation 3DGs as ours.} 
\label{table:comparison_table_online}
\centering 
\begin{tabular}{c | c | c  c  c  c  c}
\Xhline{3\arrayrulewidth}
Dataset & Method & PSNR ($\cdot$ / $\cdot$) $\uparrow$ & Storage (MB)$\downarrow$ & Train (mins)$\downarrow$ & Render (FPS)$\uparrow$ & $E_{warp}\downarrow$ \\
    
\hline
\multirow{4}{*}{MeetRoom} & StreamRF \cite{streamrf} & 26.72 / 27.53 & 5.7 & 0.170 & 10 & \cellcolor{Melon!40}{0.0100} \\
& Dynamic-3DGS* \cite{dynamic3dgs} & 27.93 / 27.58 & \cellcolor{Melon!40}{3.9} & 1.202 & \cellcolor{Magenta!30}{280} & 0.0250 \\
& 3DGStream* \cite{3dgstream} & \cellcolor{Melon!40}{30.31 / 29.37} & 4.0 & \cellcolor{Melon!40}{0.132} & 219 & 0.0110 \\
& Ours & \cellcolor{Magenta!30}{31.02 / 30.09} & \cellcolor{Magenta!30}{0.7} & \cellcolor{Magenta!30}{0.127} & \cellcolor{Melon!40}{255} & \cellcolor{Magenta!30}{0.0090} \\

\hline
\multirow{4}{*}{N3DV} & StreamRF \cite{streamrf} & 30.66 / 30.10 & 17.7 & 0.25 & 8.3 & \cellcolor{Melon!40}{0.0103} \\
& Dynamic-3DGS* \cite{dynamic3dgs} & 29.75 / 29.70 & 11.1 & 2.31 & \cellcolor{Magenta!30}{195} & 0.0278 \\
& 3DGStream* \cite{3dgstream} & \cellcolor{Melon!40}{31.84 / 30.39} & \cellcolor{Melon!40}{7.6} & \cellcolor{Melon!40}{0.23} & 145 & 0.0126 \\
& Ours & \cellcolor{Magenta!30}{32.02 / 30.78} & \cellcolor{Magenta!30}{2.1} & \cellcolor{Magenta!30}{0.22} & \cellcolor{Melon!40}{176} & \cellcolor{Magenta!30}{0.0100} \\
    
\Xhline{3\arrayrulewidth}
\end{tabular}
\end{table*}

To achieve this, we calculate the attention map, which is used to locate those $G_m$ responsible for the reconstruction of the emerging objects:
\begin{equation}        
    M_a^{i,j}=\begin{cases}
    1, \text{if} \ \mathcal{L}_{1}(\tilde{I}_d, I)^{i,j}> \tau , \\
    0, \text{else}.
    \end{cases}        
      \label{eq:attention_map}        
\end{equation}
where $i$, $j$ represent the pixel index, $\tilde{I}_d$ are rendered images using the deformed 3DGs, $\tau$ is the threshold value and empirically taken as $99$-th percentile of 
$\mathcal{L}_{1}(\tilde{I}_d, I)$. The attention map $M_a$ represents the maximum pixel error when rendered with the deformed 3DGs, indicating the mask of the emerging objects.
Similar to \cref{eq:motion}, the motion-related 3DGs $G_{new}$ responsible for the emerging object:
\begin{equation}        
    G_{new}= G_m \circ \bigcup_{i=1}^{V} (\overline{GIM}_i  \circ M_a),
      \label{eq:new_Gs}        
\end{equation}
where $\overline{GIM}$ represents the rendered GIM with the deformation applied.
The optimized spherical harmonic coefficient can be represented as:
\begin{equation}
\begin{aligned}
      \Delta sh_t = \mathcal{F}
_c (u_{G_{new}}), \\
      sh_t = sh_{t-1}+\Delta sh_t,
      \label{eq:New_object}
\end{aligned}
\end{equation}
where $u_{G_{new}}$ represents the 3DGs corresponding to emerging objects, $F_c$ is the mapping function for optimization.

\section{Experiment}\label{Results}
\subsection{Setup}
\begin{table}[!t]
\renewcommand{\arraystretch}{1.1}
\small
\caption{Quantitative comparison with offline methods on the N3DV dataset. The PSNR, storage memory, and training time are averaged over 300 frames for each scene. \colorbox{Magenta!30}{Pink} and \colorbox{Melon!40}{Brown} indicate the best and the second, respectively.}
\label{table:comparison_table_offline}
\centering 
\begin{tabular}{c | c  c  c  c }
\Xhline{3\arrayrulewidth}
\multirow{2}{*}{Method} & PSNR$\uparrow$ & Storage$\downarrow$ & Train$\downarrow$ & Render$\uparrow$ \\
& (db) & (MB) & (mins) & (FPS) \\
    
\hline
NeRFPlayer \cite{nerfplayer} & 30.69 & 17.1 & 1.2 & 0.05 \\
HexPlane \cite{hexplane} & 31.70 & 0.8 & 2.4 & 0.21 \\
K-Planes \cite{kplanes} & 31.63 & 1.0 & 0.8 & 0.15 \\
HyperReel \cite{Hyperreel} & 31.10 & 1.2 & 1.8 & 2.00 \\
MixVoxels \cite{mixvoxels} & 30.80 & 1.7 & 0.27 & 16.7 \\
4DRotorGS \cite{4drotorgs} & 31.62 & - & \cellcolor{Magenta!30}{0.20} & \cellcolor{Magenta!30}{277} \\
4DGS \cite{4dgs} & 32.01 & 20.9 & 1.1 & 72 \\
4DGaussians \cite{4DGaussians} & 31.15 & \cellcolor{Magenta!30}{0.1} & 0.34 & 147 \\ 
Spacetime GS \cite{spacetimegs} & \cellcolor{Magenta!30}{32.05} & \cellcolor{Melon!40}{0.7} & 1.0 & 140 \\
\hline
Ours & \cellcolor{Melon!40}{32.02} & 2.1 & \cellcolor{Melon!40}{0.22} & \cellcolor{Melon!40}{176} \\
    
\Xhline{3\arrayrulewidth}
\end{tabular}
\end{table}


\textbf{Datasets.}
We conducted experiments on two widely used real-world datasets, N3DV \cite{n4dnerf} and Meet Room \cite{streamrf}. The N3DV dataset is collected from 21 cameras to capture the central scene simultaneously. The videos are taken with a resolution of $2704\times 2028$ at the rate of 30 FPS. Following the previous works \cite{3dgstream,streamrf,4dgs,4DGaussians}, we downsampled the original videos into the size of $1352\times 1014$. We chose the video from camera 0 as the testing set and the rest of the videos as the training set. 

The Meet Room dataset uses 13 cameras to capture dynamic scenes, with a resolution of $1280 \times 720$ at the rate of 30 FPS. The video from the first camera was chosen as the testing set and the videos from other cameras were training set as the previous works \cite{3dgstream,streamrf}.

\textbf{Metrics.}
In order to comprehensively evaluate the performance of our method, we calculated the PSNR in the entire image and the dynamic region. We also utilized the warping error ($E_{warp}$) to assess the flickering artifacts. $E_{warp}$ is a reference-free metric and has been widely used in video de-flickering research \cite{deflickering_1,deflickering_2,deflickering_3}:
\begin{equation}
    \begin{aligned}
    E_{\text {pair }}\left(I_{i}, I_{j}\right) =\left\|M_{i, j} \odot\left(I_{i}-\mathcal{W} (I_j) \right)\right\|_{1} \\
    E_{\text {warp }}^{T} = \frac{1}{T-1} \sum^T_{t=1}  \{E_{\text {pair}}\left(I_{t}, I_{t-1}\right)+E_{\text {pair }}\left(I_{t}, I_{0}\right) \},
    \end{aligned}
    \label{eq:e_warp}
\end{equation} 
where $I_i$, $I_j$ are two different frames, $\mathcal{W}$ is the warping function to warp frame $I_j$ to $I_i$ with optical flow, $M_{i,j}$ is the occlusion mask from frame i to frame j, and T is the number of total frames. 
Besides, we also used the training and inference time, and storage memory for evaluation.


\textbf{Implementation.}
Our experiments are implemented with the original 3DGS \cite{3dgs} and are conducted on Windows OS, with the 3090 GPU. We use the original settings of the 3DGS, except for the training iterations. The 3DGS of the starting time is trained with 15000 iterations. 
Additionally, the flow mask is extracted from a pre-trained GMflow model \cite{gmflow}. The threshold for the optical flow mask and temporal difference mask are set to 1 and 10, respectively. The kernel size of the morphological function \textit{ERODE} and \textit{DILATE} is 20. The \textit{eps} and \textit{min samples}, used in clustering algorithms are set to 2 and 10, respectively. 
The training iterations of the deformation and optimization are both 100.

\subsection{Comparison}
\textit{Quantitative comparisons.} 
To achieve dynamic novel view synthesis, the \textit{offline} methods are trained with the complete videos, while the \textit{online} methods are trained in a per-frame learning paradigm. 
For comparison, we list the quantitative results in \cref{table:comparison_table_online} and \cref{table:comparison_table_offline}.
Since the \textit{offline} methods only report their results on the N3DV dataset, \cref{table:comparison_table_offline} compares MGStream with the \textit{offline} methods on the N3DV dataset.


As shown in \cref{table:comparison_table_online}, MGStream shows improved performance over the other online methods in terms of rendering quality, training/rendering/storage efficiency and warping error, demonstrating the effectiveness and superiority of the proposed method.
In addition, we conduct comparisons between MGStream and the \textit{offline} methods, and list the results in \cref{table:comparison_table_offline}. MGStream outperforms most of the \textit{offline} methods in rendering quality, and achieves competitive results on training/rendering/storage efficiency.



\textit{Qualitative comparisons.} 
We compare MGStream with three representative streaming methods: StreamRF \cite{streamrf}, Dynamic-3DGS \cite{dynamic3dgs} and 3DGStream \cite{3dgstream}. 
\cref{fig:compare_figure} shows the rendered image on three scenes, and the red and blue boxes highlight the rendering details, and the green box calculates the heatmap, indicating the temporal variance.
It is clearly seen that MGStream outperforms the other methods on rendering quality, \textit{e.g.}, StreamRF, Dynamic-3DGS and 3DGStream synthesize blur results on the head/VR handle regions in the first row, the hat in the second row and the hand in the third row, while MGStream outputs improved rendering results.
Besides, Dynamic-3DGS and 3DGStream conduct deformation on all the 3DGs, including those responsible for the reconstruction of the static, thus causing the flickering artifacts in the rendering result, as seen in the green boxes (see more details in the supplementary videos).  
In comparison, MGStream employs motion-related 3DGs for modeling the dynamic, and vanilla 3DGs for the static, guaranteeing temporal consistency.

\subsection{Ablation Study}
\textbf{Effect of the MGStream Architecture.}
To analyze the effects of the various MGStream modules, we conduct experiments on various architectural configurations, as shown in \cref{table:Ablation_architecture}. 
The \textit{OFM} represents the motion mask obtained from optical flow along, and \textit{TDM} indicates the motion mask getting from the temporal difference strategy. The \textit{CH} is the method that has applied Convex Hull Algorithms, and \textit{CA} indicates the method has used Clustering Algorithms. The \textit{OP} represents the method that has conducted the optimization phase.
\cref{table:Ablation_architecture} shows that our complete architecture achieves the best rendering performance.
Note that although the experiment without \textit{CH} achieves a relatively low warping error, the rendering performance decays substantially in the dynamic region due to the inability to move 3DGs inside the objects. 
In addition, using \textit{CH} alone not only enlarges the storage consumption and warping error but also decreases the rendering quality.
Further assessment results are included in the supplementary material.


\textbf{Effect of the \textit{Optimization}.}
To explore the impact of various training strategies, we conduct several experiments with different optimizing parameters.
As shown in \cref{table:Ablation_NOL}, the \textit{Deformation} and \textit{Optimization} represent the optimization parameters at respective stages. In specific, $\Delta u$, $\Delta r$, $\Delta c$, $\Delta o$ and $\Delta s$ represent the 3D position, rotation, color, occupancy, and scale of the motion-related $G_m$ changed in adjacent frames.
As shown in \cref{table:Ablation_NOL}, jointly training the $\Delta u$, $\Delta r$, $\Delta c$ gives the best rendering results but has the largest storage consumption and most serious flickering artifacts.
Only performing the deformation phase achieves the least memory consumption and the smallest $E_{warp}$ values. However, its rendering performance significantly decreases with a 1 dB drop. 
We choose the third setting, only optimizing color parameters in the optimization phase, for its balanced performance with probable rendering quality and storage consumption. 

\textbf{Effect of the Training Iterations.}
To search the probable iteration setting of two phases for a balanced performance according to the training efficiency and rendering performance, we conduct experiments on various numbers of iterations of \textit{Deformation} and \textit{Optimization}, respectively.
As shown in \cref{table:Ablation_iters}, setting the iteration number as 100 for both the deformation and Optimization phases provides a competitive rendering performance. Although the performance is slightly improved with more iterations, the training time is also increased.
Considering the rendering results and training efficiency, we set the iteration number as 100 for both the Deformation and Optimization phases.

\begin{table}[!t]
\caption{Ablation Study on Architecture. \textit{OFM}: Optical Flow Mask, \textit{TDM}: Temporal Difference Mask, \textit{CH}: Convex Hull, \textit{CA}: Clustering Algorithms, \textit{OP}: Optimization phase. \colorbox{Magenta!30}{Pink} indicates the setting in this paper. Best performances are highlighted in bold.} 
\vspace{-0.2em}
\label{table:Ablation_architecture}
\centering 
\footnotesize
\begin{tabular}{c | c | c | c | c | c | c | c}
\Xhline{3\arrayrulewidth}
OFM & TDM & CH & CA & OP & PSNR$\uparrow$ & Storage$\downarrow$ & Ewarp$\downarrow$\\
\hline

\textcolor{green}{\checkmark} & \textcolor{red}{\ding{55}} & \textcolor{red}{\ding{55}} & \textcolor{red}{\ding{55}} & \textcolor{red}{\ding{55}} & 32.26 & \textbf{0.66} & \textbf{0.0090} \\
\textcolor{green}{\checkmark} & \textcolor{green}{\checkmark} & \textcolor{red}{\ding{55}} & \textcolor{red}{\ding{55}} & \textcolor{red}{\ding{55}} & 32.92 & 0.67 & 0.0091 \\
\textcolor{green}{\checkmark} & \textcolor{green}{\checkmark} & \textcolor{green}{\checkmark} & \textcolor{red}{\ding{55}} & \textcolor{red}{\ding{55}} & 33.43 & 2.53 & 0.0115 \\
\textcolor{green}{\checkmark} & \textcolor{green}{\checkmark} & \textcolor{green}{\checkmark} & \textcolor{green}{\checkmark} & \textcolor{red}{\ding{55}} & 33.51 & 1.36 & 0.0102 \\

\cellcolor{Magenta!30}\textcolor{green}{\checkmark} & \cellcolor{Magenta!30}\textcolor{green}{\checkmark} & \cellcolor{Magenta!30}\textcolor{green}{\checkmark} & \cellcolor{Magenta!30}\textcolor{green}{\checkmark} & \cellcolor{Magenta!30}\textcolor{green}{\checkmark} & \cellcolor{Magenta!30}\textbf{34.31} & \cellcolor{Magenta!30} 1.80 & \cellcolor{Magenta!30} 0.0104 \\
    
\Xhline{3\arrayrulewidth}
\end{tabular}
\vspace{-0.3em}
\end{table}

\begin{table}[!t]
\caption{Ablation Study on \textit{Optimization}. \colorbox{Magenta!30}{Pink} indicates the setting in this paper. Best performances are highlighted in bold.} 
\vspace{-0.2em}
\label{table:Ablation_NOL}
\centering 
\footnotesize
\begin{tabular}{c | c | c | c | c }
\Xhline{3\arrayrulewidth}
Deformation & Optimization & PSNR$\uparrow$ & Storage$\downarrow$ & Ewarp$\downarrow$\\
\hline
\multicolumn{2}{c|}{$\Delta u$, $\Delta r$, $\Delta c$}  & \textbf{34.46} & 2.75 & 0.0162 \\

\hline

$\Delta u$, $\Delta r$ & - & 33.51 & \textbf{1.36} & \textbf{0.0102} \\
\cellcolor{Magenta!30} $\Delta u$, $\Delta r$ & \cellcolor{Magenta!30} $\Delta c$ & \cellcolor{Magenta!30} 34.31 & \cellcolor{Magenta!30} 1.80 & \cellcolor{Magenta!30} 0.0104 \\
$\Delta u$, $\Delta r$ & $\Delta c$, $\Delta o$  & 34.33 & 1.81 & 0.0107 \\
$\Delta u$, $\Delta r$ & $\Delta c$, $\Delta o$, $\Delta s$  & 34.35 & 1.83 & 0.0104 \\

\Xhline{3\arrayrulewidth}
\end{tabular}
\vspace{-0.3em}
\end{table}

\begin{table}[!t]
\caption{Ablation Study on Training Iterations. \colorbox{Magenta!30}{Pink} indicates the setting in this paper. Best performances are highlighted in bold.} 
\vspace{-0.2em}
\label{table:Ablation_iters}
\centering 
\footnotesize
\begin{tabular}{c | c | c | c | c }
\Xhline{3\arrayrulewidth}
Deformation & Optimization & PSNR$\uparrow$ & Storage$\downarrow$ & Ewarp$\downarrow$\\
\hline

25 & 100 & 32.63 & \textbf{1.75} & 0.0115 \\
50 & 100 & 33.67 & 1.79 & 0.0107 \\

\cellcolor{Magenta!30} 100 & \cellcolor{Magenta!30} 100 & \cellcolor{Magenta!30} 34.31 & \cellcolor{Magenta!30} 1.80 & \cellcolor{Magenta!30} 0.0104 \\
200 & 100 & 34.43 & 1.80 & \textbf{0.0103} \\
300 & 100 & \textbf{34.47} & 1.80 & \textbf{0.0103} \\
100 & 200 & 34.33 & 1.80 & 0.0104 \\
200 & 200 & 34.45 & 1.80 & 0.0104 \\

\Xhline{3\arrayrulewidth}
\end{tabular}
\vspace{-0.6em}
\end{table}


\section{Conclusion}\label{Conclusion}
This paper presents MGStream for streamable dynamic scene reconstruction. 
MGStream focuses on the motion-related 3DGs which are closely related to the dynamic, thus contributing to storage efficiency and temporal consistency.
To achieve this, MGStream employs the motion mask as well as the clustering-based convex hull algorithm to find the motion-related 3DGs.
The deformation and optimization are applied on the motion-related 3DGs for modeling the dynamic including the emerging objects.
The experimental results conducted on real-world datasets have demonstrated that the proposed MGStream can attain state-of-the-art (SOTA) performance in aspects such as rendering quality, training/storage efficiency, and temporal consistency.

\setcounter{section}{0}
\renewcommand\thesection{\Alph{section}}

{
    \small
    \bibliographystyle{ieeenat_fullname}
    \bibliography{main}
}

\end{document}